\def\paperID{623}
\def\confName{CVPR}
\def\confYear{2025}

\def\authorBlock{
    Junying Wang$^{1\dagger}$ \qquad
    Jingyuan Liu$^2$ \qquad
    Xin Sun$^2$ \qquad
    Krishna Kumar Singh$^2$ \qquad
    Zhixin Shu$^2$ \\
    He Zhang$^2$ \qquad
    Jimei Yang$^3$ \qquad
    Nanxuan Zhao$^2$ \qquad
    Tuanfeng Y. Wang$^2$ \qquad
    Simon S. Chen$^2$ \\
    Ulrich Neumann$^1$ \qquad
    Jae Shin Yoon$^2$ \\
    \\
    $^1$University of Southern California \qquad 
    $^2$Adobe Research \qquad 
    $^3$Runway 
}

\newif\ifreview 
\newif\ifarxiv \newcommand{\arxiv}{\arxivtrue}
\newif\ifcamera 
\newif\ifrebuttal 

\newcommand{\cmark}{\ding{51}}%
\newcommand{\xmark}{\ding{55}}%

\arxiv

\pdfoutput=1
\documentclass[10pt,twocolumn,letterpaper]{article}
\ifreview \usepackage[review]{cvpr} \fi
\ifarxiv \usepackage[pagenumbers]{cvpr} \fi
\ifrebuttal \usepackage[rebuttal]{cvpr} \fi
\ifcamera \usepackage{cvpr} \fi

\usepackage{xcolor}
\usepackage{enumitem}
\usepackage{xspace}
\usepackage{booktabs}
\usepackage{colortbl}
\usepackage{multirow}
\usepackage{makecell}
\usepackage{lipsum} 
\usepackage{gensymb} 
\usepackage{graphicx}
\usepackage{amssymb}
\usepackage{amsmath}
\usepackage{wrapfig}
\usepackage{float}
\usepackage{multicol}
\usepackage{appendix} 
\usepackage{algorithm}
\usepackage{algorithmic}
\usepackage{colortbl}
\usepackage{xcolor}   
\usepackage{url}  
\usepackage[accsupp]{axessibility}  

\definecolor{MyDarkRed}{rgb}{0.66, 0.16, 0.16}
\definecolor{MyDarkBlue}{rgb}{0.16, 0.16, 0.66}

\definecolor{gh}{RGB}{255, 224, 178}  
\definecolor{yh}{RGB}{200, 230, 201}

\makeatletter
\DeclareRobustCommand\onedot{\futurelet\@let@token\@onedot}
\def\@onedot{\ifx\@let@token.\else.\null\fi\xspace}

\makeatother

\usepackage[labelsep=period]{caption}
\renewcommand{\paragraph}[1]{\vspace{0.05cm}\noindent \textbf{#1 \hspace{0.2em}}}

\usepackage{xr-hyper}
\usepackage{amsmath}
\usepackage{amssymb}
\usepackage{color}
\usepackage{amsmath}
\usepackage{tabularx}
\usepackage{amssymb}
\usepackage{pifont}
\usepackage{arydshln}

\makeatletter
\newcommand*{\addFileDependency}[1]{
  \typeout{(#1)}
  \@addtofilelist{#1}
  \IfFileExists{#1}{}{\typeout{No file #1.}}
}

\makeatother

\definecolor{cvprblue}{rgb}{0.21,0.49,0.74}
\usepackage[pagebackref,breaklinks,colorlinks,citecolor=cvprblue]{hyperref}
\usepackage[capitalize]{cleveref}
\crefname{section}{Sec.}{Secs.}
\crefname{table}{Table}{Tables}
\crefname{figure}{Fig.}{Figs.}

\begin{document}
\title{Comprehensive Relighting: \\
Generalizable and Consistent Monocular Human Relighting and Harmonization}

\author{\authorBlock}

\twocolumn[{
\maketitle
\begin{center}
    \captionsetup{type=figure}
    \vspace{-4mm}
    \includegraphics[width=0.99\textwidth]{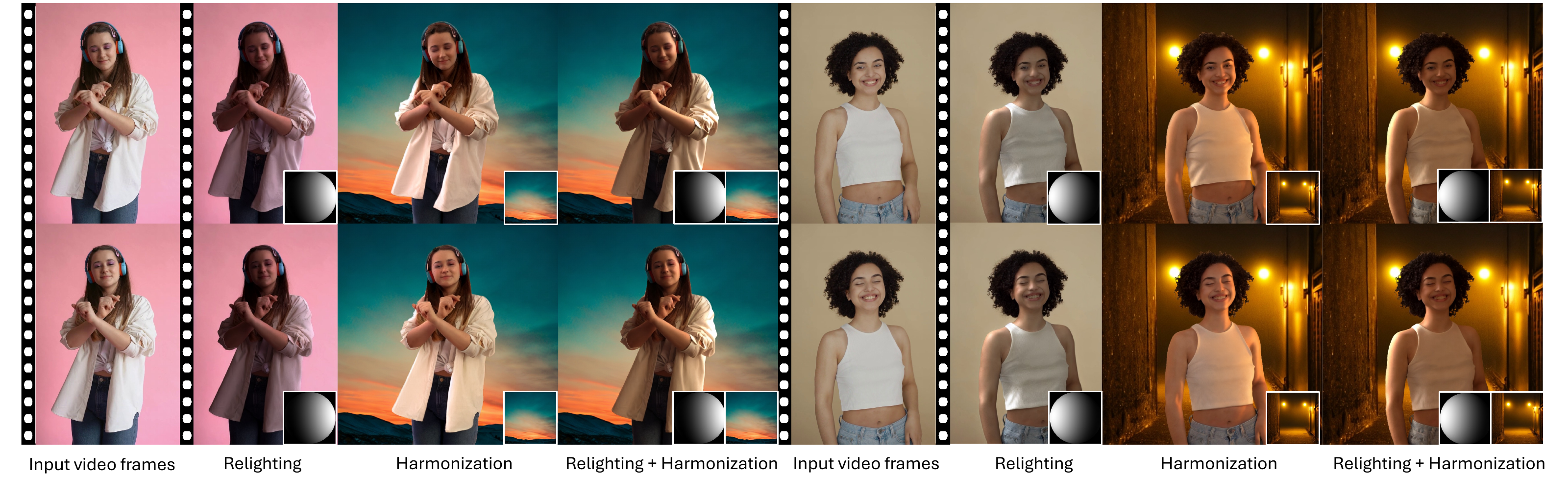}
    \captionof{figure}{We introduce Comprehensive Relighting, a generalizable and consistent model for relighting and harmonization, which controls the lighting property from a single image or video of humans with arbitrary body parts. Given target lighting coefficients, \textit{e.g.}, Spherical harmonics (second), background scenes (third), or their combination (fourth), our model performs consistent and harmonized relighting.}
    \label{fig:teaser}
\end{center}
}]

\begin{abstract}
    This paper introduces Comprehensive Relighting, the first all-in-one approach that can both control and harmonize the lighting from an image or video of humans with arbitrary body parts from any scene. Building such a generalizable model is extremely challenging due to the lack of dataset, restricting existing image-based relighting models to a specific scenario (\textit{e.g.}, face or static human). 
    \footnotetext[1]{$^\dagger$This work is partially done during an internship at Adobe Research.}
    To address this challenge, we repurpose a pre-trained diffusion model as a general image prior and jointly model the human relighting and background harmonization in the coarse-to-fine framework.
    To further enhance the temporal coherence of the relighting, we introduce an unsupervised temporal lighting model that learns the lighting cycle consistency from many real-world videos without any ground truth.
    In inference time, our temporal lighting module is combined with the diffusion models through the spatio-temporal feature blending algorithms without extra training; and we apply a new guided refinement as a post-processing to preserve the high-frequency details from the input image.
    In the experiments, Comprehensive Relighting shows a strong generalizability and lighting temporal coherence, outperforming existing image-based human relighting and harmonization methods. More demo results are available on our project page: \url{https://junyingw.github.io/paper/relighting}.
    \vspace{-6mm}    
\end{abstract}

\section{Introduction}
\label{sec:intro}

\begin{figure}[t]
\centering
\includegraphics[width=\linewidth]{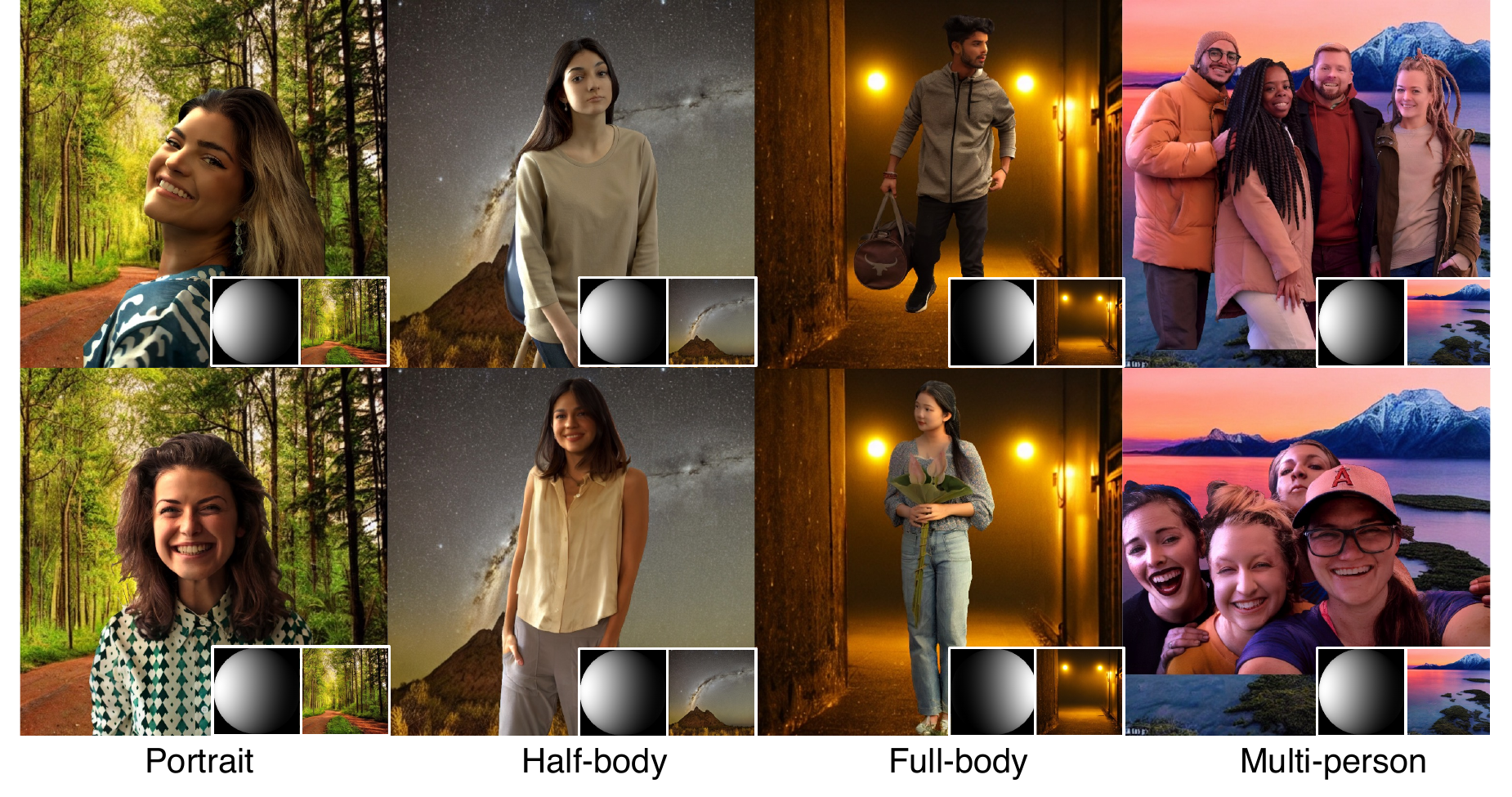}
\vspace{-5mm}
\caption{Our model generalizes to various body parts (portrait, half-body, full-body, multiperson) for relighting and harmonization, with lighting control variables shown in the insets.}
\vspace{-6mm}
\label{fig:our_test}
\end{figure}

Light is the key component that determines how a person looks, which is often defined by a specific time and space, where revisiting such a unique moment gives us the opportunity to experience authentic telepresence sensations. 
In this paper, we introduce a generalizable human relighting model that can control the lighting from an image or video of humans with arbitrary body parts (Fig.~\ref{fig:our_test}), which are well-harmonized with a conditioning space (\textit{i.e.,} background image) as shown in Fig.~\ref{fig:teaser}.

\begin{figure}[t]
    \centering
    \includegraphics[width=\columnwidth]{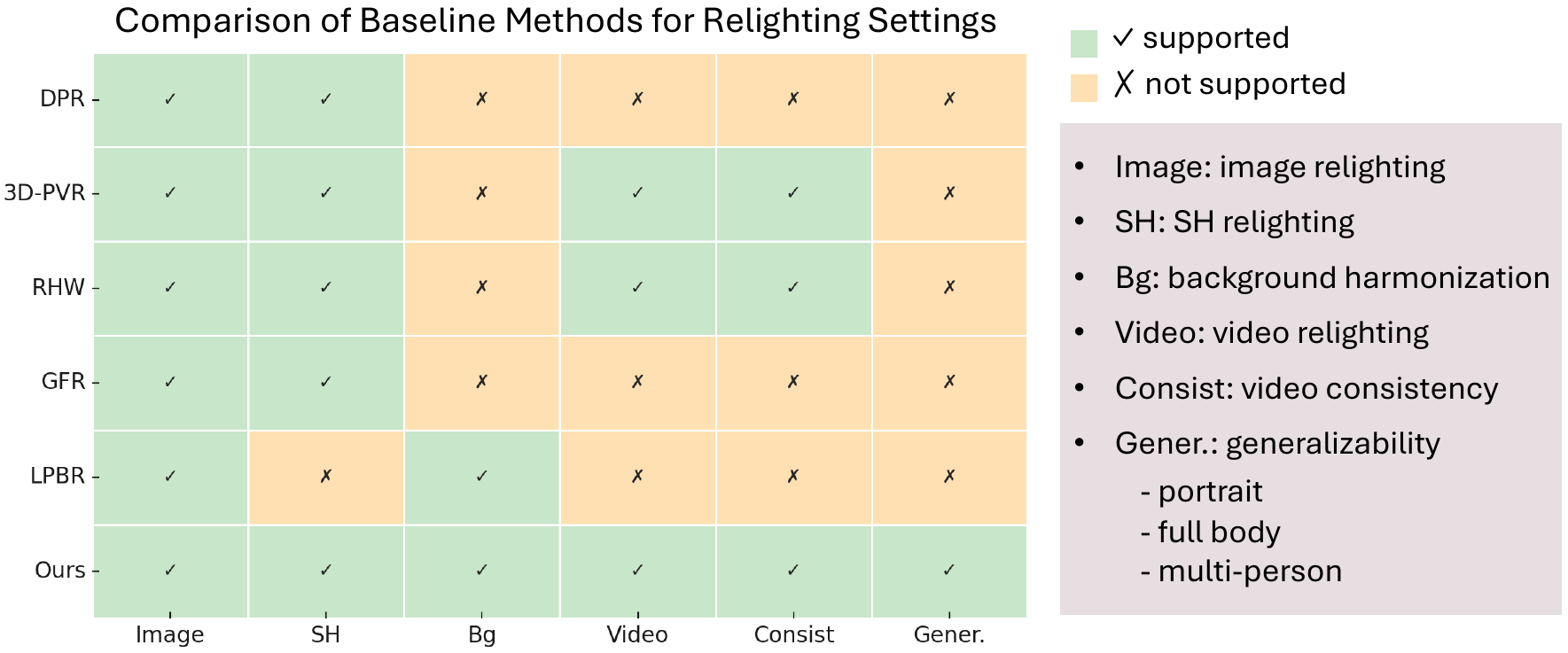}
    \caption{Comparison of various baseline methods for relighting settings and functionalities.}
    \label{fig:table1}
    \vspace{-5mm}
\end{figure}

As shown in Fig.~\ref{fig:table1}, existing image-based relighting methods face two main problems, lack of generalizability and controllability.  
First, they are designed for a specific scenario, \textit{e.g.}, only for face illumination or static humans~\cite{zhang2021neural,zhou2019deep,pandey2021total} mainly due to the scarcity of large-scale relighting datasets: Precise acquisition of the appearance of a static person under assorted lighting conditions requires specialized setups such as LightStage~\cite{pandey2021total,ji2022geometry} or expensive graphics simulation~\cite{lagunas2021single}, which are not scalable, particularly for video contents. Learning from such limited data induces weak generalization of the model to diverse scenes. 
Second, most relighting algorithms struggle to effectively model multiple light sources. Typically, these algorithms are restricted to a single lighting control from either background image (\textit{e.g.} high dynamic range lighting environment map~\cite{ji2022geometry}) or target lighting parameters (\textit{e.g.} Spherical harmonics~\cite{kanamori2019relighting}). 
These problems inhibit the production-level application that requires general use cases.

To overcome these challenges, we propose an effective approach to achieve all-in-one relighting by utilizing a diffusion model---a general image prior that learns massive visual data with diverse lighting conditions; and repurpose this prior specialized for a human relighting and harmonization in a coarse-to-fine framework:
A pretrained latent diffusion model~\cite{rombach2022high} learns from limited datasets of \footnote{To the best of our knowledge, there exists no public ground truth data for video relighting of dynamic humans, and therefore, fine-tuning an existing video diffusion model (e.g., \cite{ho2022video}) is not a readily available option.}static humans to jointly perform the fine-grained relighting and harmonization from two multi-modal lighting variables: the coarse shading estimate and conditioning background scene. 
The coarse shading estimate can be ``computed'' from conditioning lighting parameters (\textit{i.e.,} Spherical harmonics) without a neural network, and therefore, it is generalizable. 
The diffusion model is required to learn only the residual portion (\textit{e.g.,} fine self-occluded shadow), which is more generalizable than direct relighting (\textit{e.g.,} \cite{zhou2019deep}).
The pre-trained image prior helps with understanding the properties of the general background scenes, enabling the generalization of the background harmonization.

Our coarse-to-fine framework, however, still introduces problems: the diffusion model that learns to generate an image without temporal context produces significant temporal artifacts (\textit{e.g.}, sudden changes of lighting distribution even for consecutive frames). 
We address this problem by introducing an unsupervised temporal lighting model that learns from many real videos without any ground-truth data to enforce the temporal lighting consistency over frames. This temporal module learns the unsupervised cycle consistency between the relit and many real videos to predict the future lighting distribution from the past, which can be directly combined with our coarse-to-fine relighting components without extra training.

In inference time, we further push the temporal coherence of the relighting and harmonization by formulating a recurrent prediction pipeline. The generation at the current time step is conditioned onto the temporal module for the one at the next time.
The features from the lighting and temporal control modules are spatially and temporally blended to enforce temporal coherence while improving the structure of the lighting distribution.
Finally, our guided refinement module enhances the quality of the generated image in a way that preserves the original high-frequency details of the input image.

In the experiments, Comprehensive Relighting demonstrates high-quality relighting and background harmonization results with strong temporal coherence across the lighting, background, and pose changes.
It also shows strong generalizability to any unconstrained scenes regardless of body parts, views, and poses, outperforming existing relighting and background harmonization methods.

Our contributions include: (1) To the best of our knowledge, the first approach for joint modeling of relighting and background harmonization; (2) a novel coarse-to-fine framework that enables comprehensive generalization by only learning from limited synthetic and lab-controlled data; (3) unsupervised temporal modeling with lighting cycle consistency from many unconstrained real videos; (4) an effective inference algorithm with spatio-temporal feature blending and guided refinement.

\section{Related Work}
\label{sec:related_works}
\label{sec:related}
\paragraph{Image-based Human Relighting} 
While high-quality relighting often requires resource-intensive setups, recent efforts in mobile device relighting aim to address single-image scenarios. Total Relighting \cite{pandey2021total} achieves photorealistic effects by leveraging detailed normal maps and albedo as priors. However, it relies on an HDR environment map that accurately matches the scene’s real-world lighting, which is not always readily available, particularly in flexible and personalized relighting scenarios. \cite{sun2019single, ramamoorthi2001relationship} estimate and adjust the Spherical harmonics parameters from images, and others use one-light-at-a-time (OLAT) captures to generate relighting data~\cite{tewari2021monocular} or estimate reflectance fields \cite{sun2019single}. 

Recent works have focused on diverse relighting scenarios including portrait~\cite{sun2019single, futschik2023controllable, zhou2019deep, DPR, pandey2021total, liu2021relighting, zehni2021joint, hou2021towards, hou2022face, kim2024switchlight}, full-body~\cite{tajimaPG21, lagunas2021single, ji2022geometry, tajima2021relighting}, and object relighting~\cite{bi2020deep, xu2019deep, xu2018deep}. Most of these works rely on decomposing an image into its intrinsic components, \textit{i.e.,} albedo, normal, and lighting, and therefore, the accuracy of this decomposition directly impacts the quality of the final relighting.
While some works \cite{ji2022geometry, zhou2023relightable, zheng2023learning} use single-image geometry reconstruction for traditional and neural rendering, reconstruction errors are often propagated to relighting results. Tajima et al. \cite{tajima2021relighting} tackled domain adaptation with a two-step relighting framework, yet noticeable texture distortion remains due to limited model generalizability. Zhang \textit{et al.}~\cite{zhang2023fdnerf} trained a 2D latent-diffusion model, allowing users to manipulate and construct face NeRFs in a zero-shot learning framework without the need for explicit 3D data.
DiFaReli~\cite{ponglertnapakorn2023difareli} utilizes DDIM (Denoising Diffusion Implicit Models~\cite{song2020denoising}) for high-fidelity face relighting. While promising, their methods are constrained by a focus on either specific body parts, limiting their applicability for generalizable.

\paragraph{Background Harmonization}
Background harmonization seeks to harmonize color, contrast, and style discrepancies between the foreground and background, ensuring composite images appear natural and cohesive. 
Many existing background harmonization methods~\cite{zhu2015learning,tsai2017deep,guo2021intrinsic,jiang2021ssh,cong2020dovenet,guo2021image,cong2022high,PIH_wang2023semi,INR,PCT_Guerreiro_2023_CVPR} formulate this problem as image-to-image translation work where a neural network translate an unharmonized foreground image to the harmonized one in the context of the conditioning background image.
Recently, Relightful Harmonization~\cite{ren2023relightful} introduced methods that harmonize both image style and lighting to match the background scene, and IC-Light~\cite{zhang2025scaling} further enables flexible illumination control via image diffusion models, guided by text descriptions or background images.
While promising, these methods lack explicit lighting control, restricting general applicability and consistent video relighting. Additionally, some (\cite{ren2023relightful}) are limited to specific body parts (\textit{e.g.}, portraits), with constrained support for full-body and multi-person scenarios.

\paragraph{Video Relighting}
Neural Radiance Fields (NeRF)~\cite{mildenhall2021nerf} based methods enable novel view synthesis under varying lighting conditions in videos \cite{zhou2023relightable, toschi2023relight, shuai2022novel, wang2023learning}. Zhang~\textit{et al.}~\cite{zhang2021neural} achieve portrait video relighting under dynamic illuminations, while Choi~\textit{et al.}~\cite{choi2024personalized} ensure temporally consistent relit videos. 3D-PVR~\cite{cai2024real} present a 3D-aware, real-time method to relight videos of talking faces. Relighting4D~\cite{chen2022relighting4d} decomposes the time-varied human body as a set of neural fields of normal, occlusion, diffuse, and specular maps. However, their rendering quality are highly reliant on geometry accuracy. 
Some works~\cite{ponglertnapakorn2023difareli} adopt a temporal modeling scheme real-time neural video portrait relighting. ST-NeRF~\cite{zhang2021editable} controls dynamic scenes using a neural layered radiance representation that maintains spatial and temporal coherence. Li \textit{et al.}~\cite{li2014free} employed multi-view reconstruction for free-viewpoint relighting videos under general illumination. Richardt \textit{et al.}~\cite{richardt2012coherent} use RGBZ video cameras for video effects, including relighting, but rely on multi-view reconstruction from specialized hardware, making the process highly complex and costly.

\begin{figure*}[t]
\centering
\includegraphics[width=1\textwidth]{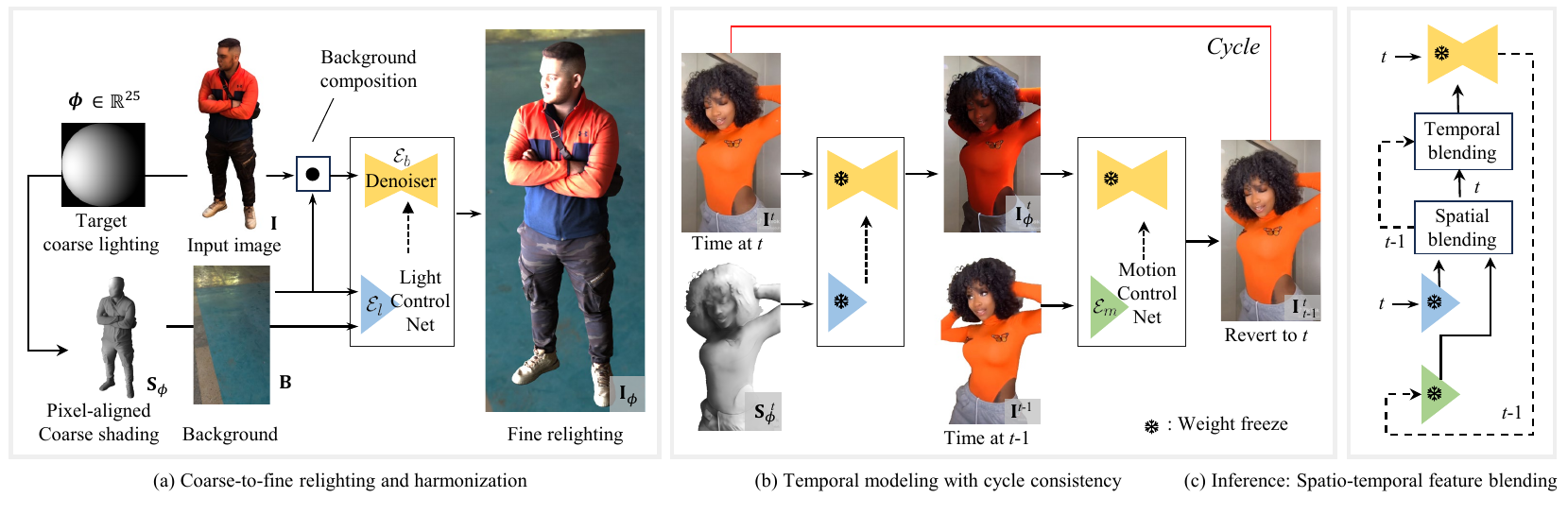}
\vspace{-6mm}
\caption{\textbf{System overview.} (a) Given an input image of humans with coarse lighting and background image, our diffusion model generates the relit images harmonized with background scenes (Sec.~\ref{sec2}). (b) The external temporal modules learn the temporal cycle consistency from many real-world videos to construct temporal lighting features (Sec.~\ref{sec3}). (c) In inference time, we blend the features from lighting and temporal modules spatially and temporally to enable coherent and generalizable human relighting (Sec.~\ref{inference}).} 
\label{fig:overview}        
\vspace{-3mm}
\end{figure*}
\section{Method}%
\label{sec:method}
We develop a generalizable and consistent human relighting and harmonization framework using a diffusion model. 
Fig.~\ref{fig:overview} illustrates the overview of our framework. 
Given an input image of humans and control lighting variables, including coarse shading and background image, a diffusion model learns to generate a fine-grained image of the humans under the target lighting, also harmonized with the background (Sec.~\ref{sec2}). 
An external temporal lighting module, trained on real videos using unsupervised temporal cycle consistency, is integrated into the diffusion model to enhance temporal lighting coherence (Sec.~\ref{sec3}).
In inference time, we blend the features between the lighting and temporal control modules over time to ensure the relighting results are accurate and temporally coherent, and we apply a guided refinement to prevent the loss of high-frequency details during the denoising process (Sec.~\ref{inference}). 

\subsection{Background: Image-based Relighting}\label{sec1}
Image-based relighting function can be compactly modeled with a small number of approximated basis of Spherical harmonics (SH)~\cite{schonefeld2005spherical,tajimaPG21,lagunas2021single} which describes only essential features of illumination on the surface of a 3D sphere based on the following formulation: 
\begin{eqnarray}
I(\mathbf{x}) = \rho(\mathbf{x})\cdot\sum_{l=0}^{k} \sum_{m=-l}^{l} {\phi} Y_{lm}(\mathbf{n(x)})
\label{sh_equ}
\end{eqnarray}
where $\phi \in \mathbb{R}^{(k+1)^2}$ denotes the spherical harmonics (SH) coefficients; $Y_{lm}(\mathbf{n})$ are the SH basis functions evaluated at the surface normal $\mathbf{n}$; indices $l$ and $m$ represent the band and order within each band, respectively; and $k$ is the maximum order of spherical harmonics used. While SH is computationally efficient and highly generalizable, it only captures low-frequency lighting, limiting its ability to model fine-grained, high-frequency illumination details. Additionally, SH-based global illumination neglects the context provided by background images, often resulting in relit images appearing unnatural when composited with different backgrounds. To overcome these limitations, we propose a coarse-to-fine human relighting and harmonization framework that leverages the strong image prior available from a pre-trained text-to-image diffusion model.
\subsection{Coarse-to-Fine Relighting and Harmonization}\label{sec2}
We generate a fine-grained relit image of a person conditioned on a coarse lighting representation:
\begin{eqnarray}
\mathcal{E}(\mathbf{I};\mathbf{S}_{\phi})=\boldsymbol{z}, \ \ \ \mathcal{D}(\boldsymbol{z})= \mathbf{I}_{\phi}   
\label{eq2}
\end{eqnarray}
where $\mathcal{E}$ is an encoder that generates the latent lighting features, $\boldsymbol{z}$ as a function of the input image $\mathbf{I}\in\mathbb{R}^{w\times h\times 3}$ and a small number of global lighting parameters $\boldsymbol{\phi}\in\mathbb{R}^{n}$ (\textit{i.e.}, Spherical harmonics where $n=25$); and $\mathcal{D}$ is the decoder that generates fine-grained relit image $\mathbf{I}_{\phi}\in\mathbb{R}^{w\times h\times 3}$ from $\boldsymbol{z}$. $\mathbf{S}_{\phi}\in\mathbb{R}^{w\times h}$ is the pixel-aligned coarse shading map converted from the coarse lighting parameters $\phi$ as shown in Fig.~\ref{fig:overview}-(a).
One approach to obtaining $\mathbf{S}_{\phi}$ is directly computing it as demonstrated in Eq.~\ref{sh_equ} where we can compute lighting intensity along with the detected surface normal map from $\mathbf{I}$ given spherical harmonics coefficients $\boldsymbol{\phi}$.
Another way, is to use a neural network to convert the surface normal to $\mathbf{S}_{\phi}$ as a condition of spherical harmonics coefficients ${\boldsymbol\phi}$, \textit{i.e.,} ${\mathbf{S}}_{\phi}\leftarrow f(\mathbf{N};\boldsymbol{\phi})$ where $f$ is the neural shading function which maps the surface normal $\mathbf{N}$ and $\boldsymbol{\phi}$ to the coarse shading. While both approaches are highly generalizable, our experiments indicate that the latter method achieves improved accuracy and better noise correction. Please refer to the Supple. documents for more details about the coarse shading estimation.

To jointly model the relighting and background harmonization, we further condition a target background image $\mathbf{B}\in\mathbb{R}^{w\times h\times 3}$ as the additional lighting sources:
\begin{eqnarray}
\mathcal{E}(\mathbf{I};\{{\mathbf{S}}_{\phi}, \mathbf{B}\})=\boldsymbol{z},  \ \ \ \mathcal{D}(\boldsymbol{z})= \mathbf{I}_{\phi}
\label{eq4}
\end{eqnarray}
where the lighting encoder $\mathcal{E}$ learns to capture the intensity and direction of light from $\mathbf{S}_{\phi}$ while capturing ambient environment lighting and color distribution from $\mathbf{B}$. This enables $\mathcal{D}$ to achieve complete relighting in scenarios involving new target lighting, background, or both. In the subsection, we enable $(\mathcal{D}\circ\mathcal{E})$ using a latent diffusion model.

\subsubsection{Fine-grained Relighting Diffusion Model}\label{sec:diffusion}
We enable the fine-grained image relighting using a diffusion model as illustrated in Fig.~\ref{fig:overview}-(a).
Our encoder $\mathcal{E}$ in Eq.~\ref{eq4}, in practice, is formulated as the composition of two encoders:
\begin{eqnarray}
\mathcal{E}\rightarrow \mathcal{E}_{\rm b}(\mathbf{I}; \mathcal{E}_{\rm l}(\{{\mathbf{S}}_{\phi},\mathbf{B}\}))=\boldsymbol{z},\ \ \mathcal{D}(\boldsymbol{z})= \mathbf{I}_{\phi} 
\label{eq6}
\end{eqnarray}
where $\mathcal{E}_{\rm l}$ encodes lighting control variables, $\{{\mathbf{S}}_{\phi}, \mathbf{B}\}$, and $\mathcal{E}_{\rm b}$ encodes the conditional variable $\mathbf{I}$ whose visual properties, \textit{e.g.}, semantics and identity, should be preserved in the output along with the controls from $\mathcal{E}_{\rm l}$.
For $\mathcal{E}_{\rm b}$, we use the base latent diffusion model~\cite{brooks2023instructpix2pix, rombach2021highresolution} pre-trained for text-to-image generation task, and for $\mathcal{E}_{\rm l}$, we use ControlNet~\cite{zhang2023adding} (termed as Light ControlNet in Fig.~\ref{fig:overview}-(a)). While noise, texts, and timestep variables are also conditioned on $\mathcal{E}_{\rm b}$ to fit the modality of the latent diffusion model, they are not described in the equations and figures for conciseness. 
To impose the foreground attention on the lighting control, a portrait mask is also conditioned to $\mathcal{E}_{l}$:
\begin{eqnarray}
\mathcal{E}_{\rm b}(\mathbf{I}; \mathcal{E}_{\rm l}(\{{\mathbf{S}}_{\phi},\mathbf{B}\};\mathbf{M}))=\boldsymbol{z},\ \  \mathcal{D}(\boldsymbol{z})= \mathbf{I}_{\phi}
\label{eq7}
\end{eqnarray}
where $\mathbf{M}\in\{0,1\}^{w\times h}$ is the foreground binary mask.
In training time, the diffusion model learns to directly change the lighting distribution of the input images (without inverse rendering techniques) as the condition of coarse lighting estimate by minimizing the latent distance of the noisy relit image with the clean one from the ground truth in the overall forward and background denoising steps. Following the findings from an existing harmonization work~\cite{ren2023relightful}, we use the composite image between the input image and conditioning background image as $\mathbf{I}$. We randomly drop background $\mathbf{B}$ or randomly set the coarse shading $\mathbf{S}_{\phi}$ as binary mask such that the diffusion model learns the control the lighting from $\mathbf{B}$, $\mathbf{S}_{\phi}$, or both $\{\mathbf{B}, \mathbf{S}_{\phi}\}$.

\subsection{Unsupervised Add-on Temporal Modeling}\label{sec3}
Our coarse-to-fine relighting framework ($\mathcal{D}\circ\mathcal{E}_{\rm b}\circ\mathcal{E}_{\rm l}$) is trained only on individual images, inherently missing temporal context, \textit{e.g.}, how a point on a human's surface will radiate from a specific viewpoint under the continuous pose, view, and illumination changes, leading to temporal artifacts such as flickering.
We model such temporal context by designing an external add-on temporal lighting module $\mathcal{E}_{m}$ that can be combined, in inference time, with the relighting framework without extra training:
\begin{eqnarray}
\mathcal{D}\circ\mathcal{E}_{\rm b}\circ\mathcal{E}_{\rm l} \rightarrow \mathcal{D}\circ\mathcal{E}_{\rm b}\circ(\mathcal{E}_{\rm l}\times \mathcal{E}_{\rm m})
\label{eq7}
\end{eqnarray}
To enable this, our temporal module is designed to regress the relit image from the previous time instance, \textit{i.e.,} $\mathbf{I}_{\phi}^{t-1}$, to the latent lighting distribution whose space is shared with our base relighting models. Therefore, our decoder can generate temporally coherent relit images in the current time in an auto-regressive way:
\begin{eqnarray}
\mathcal{D}(\mathcal{E}_{\rm b}(\mathbf{I}^{t};\mathcal{E}_{\rm m}(\mathbf{I}_{\phi}^{t-1})))=\mathbf{I}^{t}_{t-1}.
\label{eq8}
\end{eqnarray}
However, training $\mathcal{E}_{\rm m}$ with a conventional L2 loss is not possible since there exists no ground-truth video relighting data for dynamic humans. Therefore, we propose to learn $\mathcal{E}_{\rm m}$ in an unsupervised way using many real videos with lighting cycle consistency as follows:
\begin{gather}
\mathcal{D}^{*}(\mathcal{E}_{\rm b}^{*}(\mathbf{I}^{t};\mathcal{E}_{\rm l}^{*}(\{\mathbf{S}_{\phi}^{t}, \mathbf{B}^{t}\}; \mathbf{M}^{t})))=\mathbf{I}^{t}_{\phi}\ \ 
 \label{forward},\\ 
\mathcal{D}^{*}(\mathcal{E}^{*}_{\rm b}(\mathbf{I}^{t}_{\phi};\mathcal{E}_{\rm m}(\mathbf{I}^{t-1},\mathbf{M}^{t-1})))=\tilde{\mathbf{I}}^{t}_{t-1}  \label{backward}\\
\therefore \ \ \  \mathbf{I}^{t}=\tilde{\mathbf{I}}^{t}_{t-1}.\nonumber
\end{gather}
We make the hypothesis: a video sequence inherently contains temporal lighting properties whose flow can be implicitly modeled by learning to predict the lighting distribution of the future frame based on the hint from the previous frame. This involves forward and backward processes, as a cycle-training. Eq.~\ref{forward} represents the forward image relighting, \textit{i.e.,} $\mathbf{I}^{t}\rightarrow\mathbf{I}^{t}_{\phi}$, where our relighting diffusion model with lighting ControlNet $\mathcal{E}_{\rm l}$ generates the relit image at time $t$ as a condition of a novel lighting condition (where we pick random Spherical harmonics). 
Eq.~\ref{backward} reverts the relighting, \textit{i.e.}, $\tilde{\mathbf{I}}^{t}_{t-1}\leftarrow\mathbf{I}^{t}_{\phi}$ to the original input image conditioned by the original frame in the previous time step through our temporal lighting module where the mask $\mathbf{M}$ is used for foreground awareness. 
Finally, $\mathcal{E}_{m}$ learns the lighting cycle consistency in the diffusion process by minimizing the latent distance between the reverted and the original image.
We freeze the learnable weights for the functions with $*$ during training, and $\mathcal{E}_{\rm m}$ is enabled with another ControlNet~\cite{zhang2023adding} termed as Motion ControlNet in Fig.~\ref{fig:overview}-(b). 
{\renewcommand{\tabcolsep}{5.8pt}
\begin{table*}
    \footnotesize
    \centering
    \scalebox{1.02}{
    \begin{tabular}{|l||p{2.1cm}|p{2.1cm}|p{2.1cm}|p{2.3cm}|p{2.1cm}|p{2.1cm}|}
    \hline
        \multicolumn{1}{|c||}{\textbf{Method}} &
        \multicolumn{2}{c|}{\textbf{Scenario 1}} &
        \multicolumn{2}{c|}{\textbf{Scenario 2}} &
        \multicolumn{2}{c|}{\textbf{Scenario 3}} \\
    \hline
    DPR~\cite{zhou2019deep} 
        & 18.62/0.86/0.103 
        & \colorbox{gh}{32.00}/0.94/0.030 
        & 18.14/0.89/0.089 
        & \colorbox{gh}{35.29}/0.98/0.032 
        & 21.20/0.89/0.072 
        & 37.11/0.95/0.038 \\
    RHW~\cite{tajima2021relighting} 
        & 19.78/0.87/0.113 
        & 30.74/0.95/\colorbox{gh}{0.027} 
        & 20.12/0.88/0.078 
        & \colorbox{yh}{36.64}/0.98/\colorbox{yh}{0.028} 
        & 23.33/0.90/0.060 
        & \colorbox{gh}{37.53}/0.98/0.033 \\
    GFR~\cite{ji2022geometry} 
        & \cellcolor{gh}{25.59/0.91/0.089} 
        & 30.33/0.95/0.033 
        & \cellcolor{gh}{22.76/0.91/0.072} 
        & 32.36/0.98/0.036 
        & \cellcolor{gh}{25.49/0.93/0.050} 
        & 35.89/0.98/\colorbox{gh}{0.028} \\
    LPBR~\cite{ren2023relightful} & 18.19/0.86/0.090 & 31.62/0.91/0.035 & 19.96/0.88/0.084 & 27.94/0.94/0.041 & 21.42/0.88/0.053 & 33.39/0.95/0.038 \\
    Ours & \cellcolor{yh}{25.95/0.95/0.066} &
               \cellcolor{yh}{33.58/0.96/0.024} &
               \cellcolor{yh}{23.99/0.93/0.048} &
               35.18/\colorbox{yh}{0.99}/\colorbox{gh}{0.031} &
               \cellcolor{yh}{26.61/0.94/0.033} &
               \cellcolor{yh}{38.32/0.98/0.026} \\
    \hline
    \end{tabular}}
    \caption{
    Comparison with existing image-based human relighting methods on synthetic videos for fidelity and temporal consistency. Each column shows image fidelity (left) and video temporal consistency (right). Metrics are PSNR$\uparrow$ / SSIM$\uparrow$ / LPIPS$\downarrow$ for accuracy, and tPSNR$\uparrow$ / tSSIM$\uparrow$ / tLPIPS$\downarrow$ for temporal consistency. While \colorbox{yh}{green} is used for the best values, \colorbox{gh}{yellow} highlights the second-best values.}
    \vspace{-2mm}
    \label{tab:table2}
\end{table*}
}

\begin{figure*}[t]
\centering
\includegraphics[width=\linewidth]{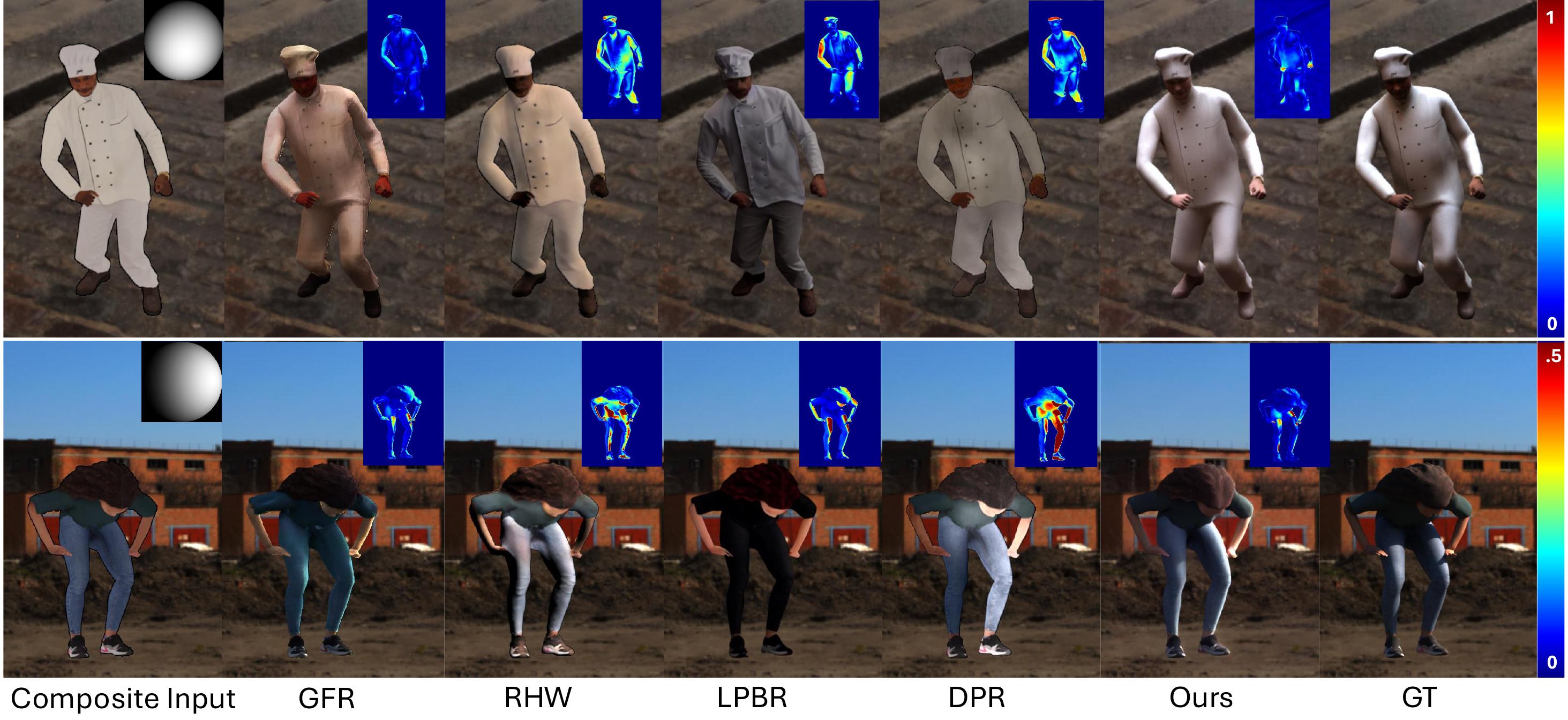}
\caption{Qualitative comparison of synthetic video frames (corresponding to Tab.~\ref{tab:table2}). From left to right: composite input with target lighting parameters (inset), our relit result, baseline methods, and normalized L2$\downarrow$ photometric error map (inset).}
\vspace{-4mm}
\label{fig:video_compare}
\end{figure*}

\subsection{Inference with Spatio-Temporal Blending and Guided Refinement}\label{inference}
Given a video or image, we perform comprehensive relighting with a spatio-temporal feature blending framework.
For $t=0$, we generate the relit image without our temporal lighting module $\mathcal{E}_{\rm m}$.
For $t>0$ (note that even for the static image, $t>0$ is possible since the lighting is time-variant), the generated relit images in the previous time step are conditioned on $\mathcal{E}_{\rm m}$, and therefore, the relighting is controlled by dual control modules, \textit{i.e.}, $\mathcal{E}_{\rm l}$ and $\mathcal{E}_{\rm m}$ by blending their features with a (0.85:0.15) ratio as described as spatial blending block in Fig.~\ref{fig:overview}-(c). 
The blended lighting features are recurrently combined with the one from the previous time step through the temporal blending block. For this, we adopt optical flow from the original input video as a temporal prior to spatially align the features from consecutive frames similar to ~\cite{Lei_2023_CVPR}; and we blend the aligned temporal features with a (0.5:0.5) ratio as in Fig.~\ref{fig:overview}-(c) to improve the temporal consistency. 

In the denoising process of the latent diffusion model, the generation often suffers from the loss of high-frequency details. Inspired by existing image restoration techniques~\cite{zhang2018residual,song2023guided}, we address this problem by applying a guided refinement to each generated image. We cast the problem of guided refinement as a guided residual prediction:
\begin{eqnarray}
\mathbf{I}_{\phi}^{\rm refine}=\mathbf{I}+g(\mathbf{I}_{\phi},\mathbf{I};\mathbf{M}),
\label{eq11}
\end{eqnarray}
where $g$ is the function that predicts the guided lighting residual. 
This residual learns to map the lighting distribution from $\mathbf{I}$ to $\mathbf{I}_{\phi}$. Here, 
$\mathbf{I}_{\phi}^{\rm refine}$ can effectively preserve the high-frequency details of the input image $\mathbf{I}$ due to the nature of residual learning~\cite{zhang2018residual,yang2017deep} that is designed to preserve the visual properties from the observation space, \textit{i.e.}, $\mathbf{I}$. We enable $g$ with a residual network~\cite{he2016deep} by learning from our relighting data with general losses for low-level vision processing, \textit{i.e,} L2 and VGG~\cite{gatys2016image}.

\vspace{-1mm}
\section{Experiments}%
\label{sec:Experiments}
\vspace{-2mm}
\noindent\textbf{Dataset.}\label{sec:dataset}
To train our coarse-to-fine relighting model, we use 100K ground-truth relighting samples, including 50K synthetic human renderings and 50K OLAT-captured images from LightStage, with random cropping augmentation. Ground-truth albedo, images, background, masks, and lighting coefficients are precomputed. Our dataset is categorized by gender, skin tone, and body part (full-body and multi-body), with each subject captured from 32 viewpoints under varying lighting conditions. For a detailed dataset breakdown, refer to the Supple.
To train our temporal control module using real data, we used 50K frames of videos from existing works~\cite{jafarian2021learning,Perazzi2016,lwb2019,li2021learn} and customized videos, where, for cyclic relighting, we randomly sample spherical harmonics lighting parameters from ground-truth data.
For validation on static humans with dynamic lighting, we generated a new testing set (\textit{i.e.,} the ground-truth relit images and associated lighting parameters) using an internal simulation and LightStage. This set includes the mixture of half-body, full-body, and multi-person scenarios, with each scenario comprising 100 frames. 
For the validation on dynamic humans (\textit{i.e.}, video), we newly create synthetic video sequences for three scenarios: Scenario 1 involves static lighting (environment map) across frames with a moving human; Scenario 2 features dynamic lighting (rotated environment map) with a static human; Scenario 3 combines dynamic lighting (rotated environment map) with a moving human. Please refer to the Supple. and demo video for more details about the testing results.
%

\begin{figure}[t]
\centering
\includegraphics[width=\linewidth]{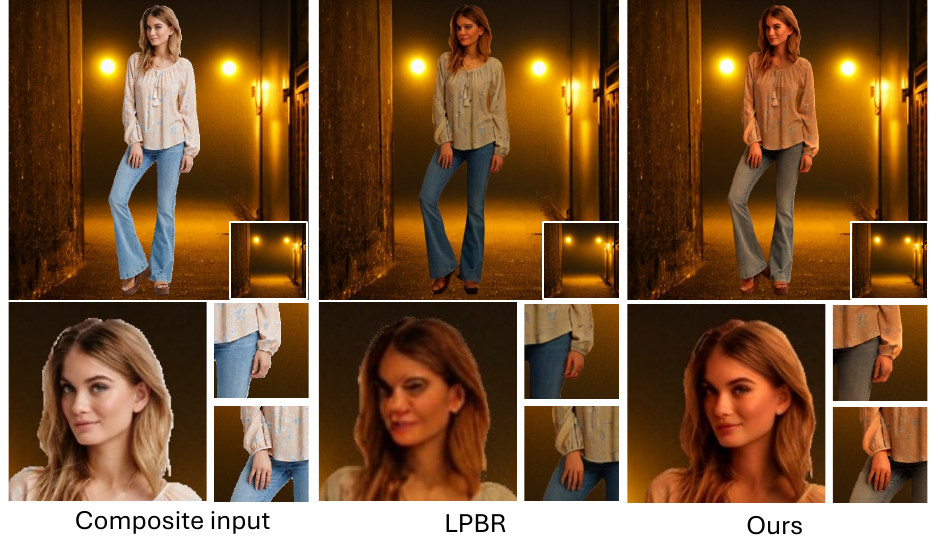}
\vspace{-6mm}
\caption{Comparison with LPBR~\cite{ren2023relightful} on DeepFashion~\cite{liu2016deepfashion} real images for background harmonization testing. The first row shows the relit output, and the second shows the magnified results.}
\vspace{-4mm}
\label{fig:comp_lpbr}
\end{figure}

\begin{figure*}[t]
\centering
\includegraphics[width=\linewidth]{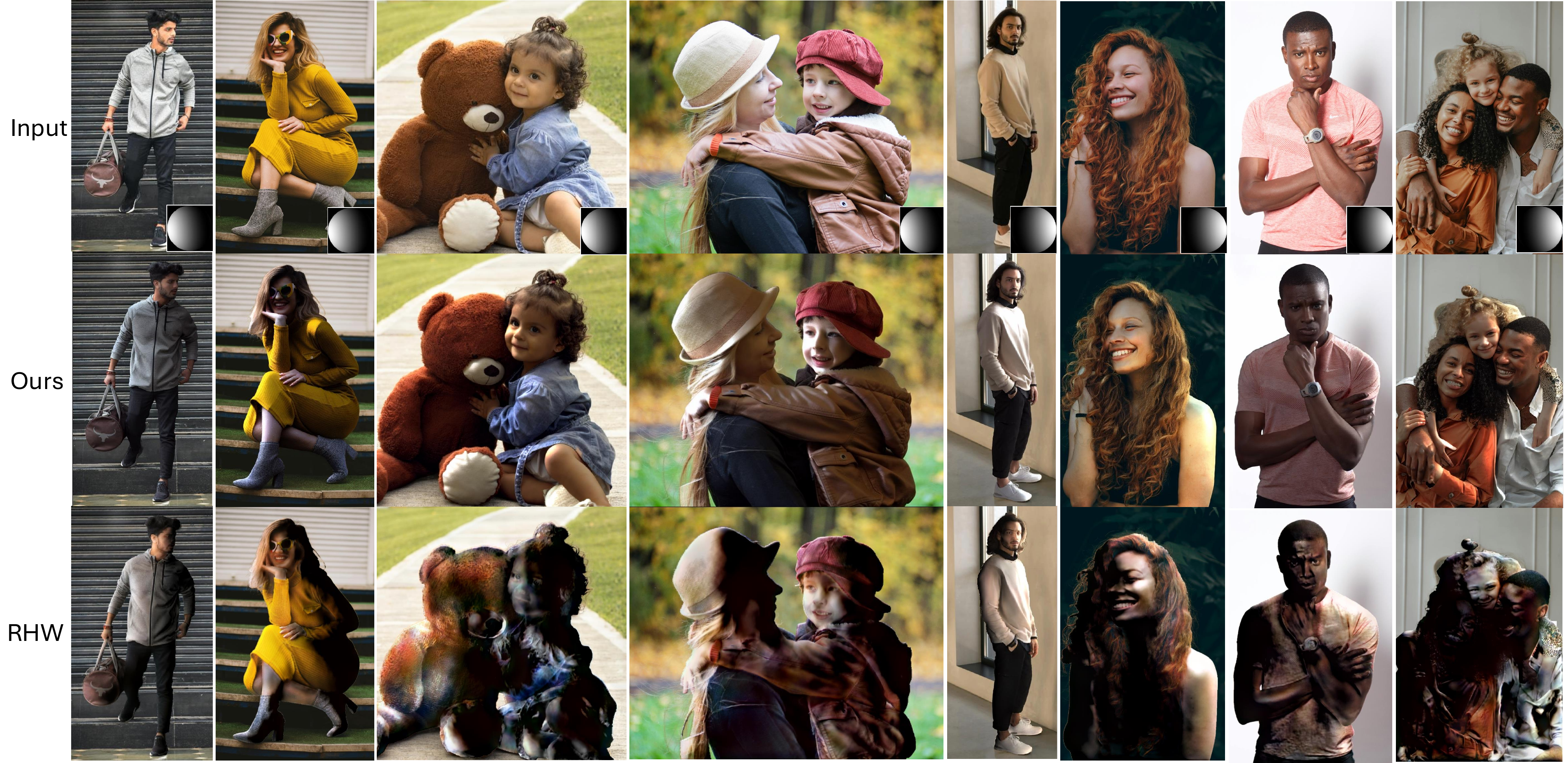}
\caption{Comparison with RHW~\cite{tajima2021relighting} on Pexels~\cite{Pexels} real images. The lighting control variables are shown as insets. While RHW produces reasonable relighting for full-body images, its quality degrades on half-body and multi-person cases.}
\label{fig:res1}
\end{figure*}

\begin{figure*}[t]
\centering
\vspace{-2mm}
\includegraphics[width=\linewidth]{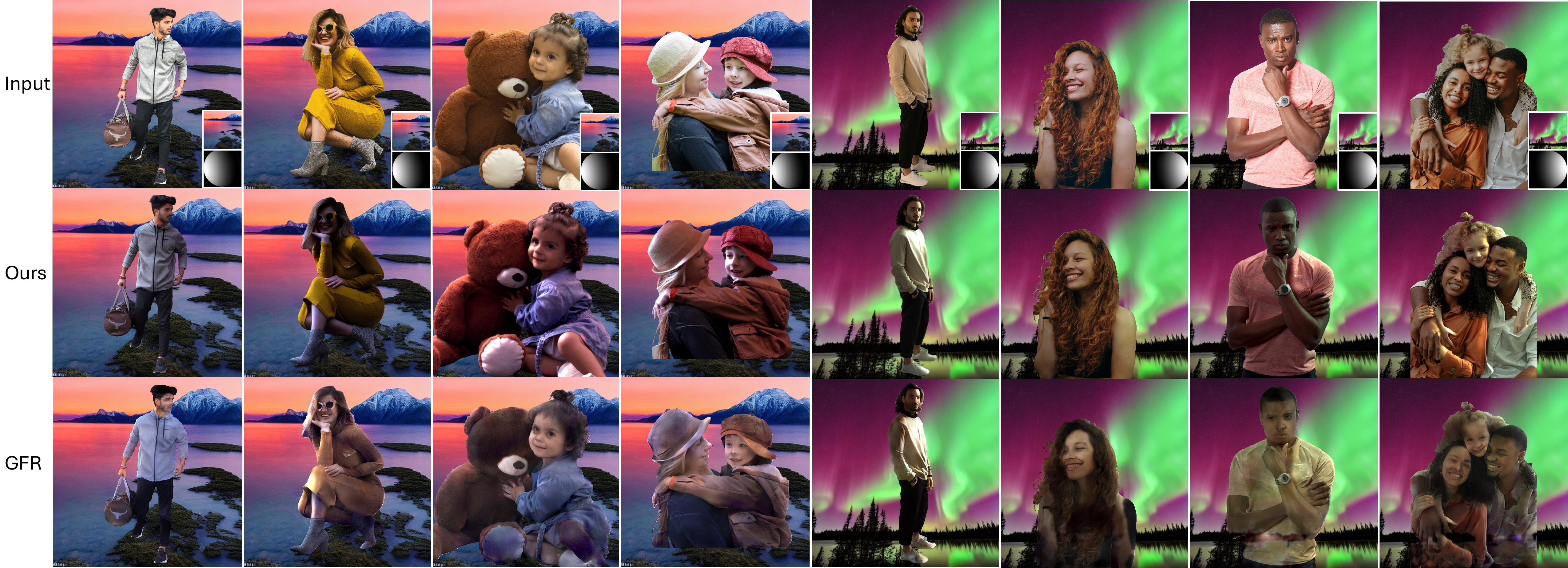}
\caption{Comparison with GFR~\cite{ji2022geometry} on Pexels~\cite{Pexels} real images. The lighting control variables are shown as insets. Limited generalizability of GFR results in reduced output quality for half-body and multi-person cases.}
\label{fig:comp_gfr}
\vspace{-5mm}
\end{figure*}

\noindent\textbf{Metrics.} 
To measure the relighting quality, we use L1 distance, reconstruction fidelity (PSNR~\cite{hore2010image}), local structure similarity (SSIM~\cite{hore2010image}), and the latent perceptual distance (LPIPS~\cite{zhang2018perceptual}), between the relighted and ground-truth.
For measuring the temporal coherence of the relighting results, we follow the same logics as TokenFlow~\cite{tokenflow2023}. This involves warping the relit image from the previous time step to the next using optical flow~\cite{teed2020raft}, and then comparing this to the current frame using the aforementioned metrics, termed t$L_1$ error, tPSNR, tSSIM, and tLPIPS.

\noindent\textbf{Baseline.}
We compare our method with existing monocular human relighting works where we chose the baselines that are applicable to general scenes (\textit{e.g.,} any part-specific information such as a 3D face model
is not a requirement): DPR~\cite{DPR}, RHW~\cite{tajima2021relighting}, and GFR~\cite{ji2022geometry}. For the harmonization baseline, we compare our model with LPBR~\cite{ren2023relightful}, a diffusion-based light-aware harmonization method. For relighting, aside from GFR, baselines use Spherical harmonics for lighting control without modeling background illumination. DPR and RHW are evaluated using their released pre-trained models. Since GFR lacks available code, we re-implemented it using our dataset, replacing HDR lighting with Spherical harmonics and background modeling to enable background harmonization in our experiments. Harmonization is directly compared with LPBR by replacing the background.

\noindent\textbf{Results.} \label{sec:compare_result}
For testing static humans under dynamic lighting, we present the quantitative comparison in Tab.~\ref{tab:breakdown}. We show average numerical evaluations on our synthetic testing dataset, categorized by portrait, full-body, and multi-person. Further validation by gender, and skin color is detailed in the Supple. Methods such as RHW and DPR have limited generalizability for both full-body and portrait relighting. They are difficult to extend to other scenarios and tend to show reduced performance when tested in different settings. While GFR performs well in our evaluation, it struggles with significant domain gaps, leading to noticeable quality degradation on real data, including distortions and color shifts (Fig.~\ref{fig:comp_gfr}). In contrast, our model exhibits strong generalizability across validation sets and real-world tests (Figs.~\ref{fig:res1}, ~\ref{fig:comp_gfr}).
Compared to the state-of-the-art background harmonization method (LPBR), our method shows the strong generalization to different body parts, and notably, LPBR often includes significant distortion when applied to the full-body images and it does not support the lighting control function as shown in Fig.~\ref{fig:comp_lpbr}.
{\renewcommand{\tabcolsep}{2.8pt} 
\begin{table}[H]
    \centering
    \vspace{-1mm}
    \scalebox{0.8}{
    \begin{tabular}{|l||c|c|c|c|}
    \hline
    \textbf{Category} & \textbf{DPR~\cite{zhou2019deep}} & \textbf{RHW~\cite{tajima2021relighting}} & \textbf{GFR~\cite{ji2022geometry}} & \textbf{Ours} \\
    \hline
    Portrait & 17.74 / 0.87 & 15.75 / 0.82 & 17.71 / 0.86 & \cellcolor{yh}{23.04 / 0.90}  \\
    Full-body & 27.62 / 0.96 & 27.73 / 0.95 & 29.51 / 0.95 & \cellcolor{yh}{30.81 / 0.97} \\
    Multi-person & 25.70 / 0.95 & 25.69 / 0.95 & 29.35 / \colorbox{yh}{0.97} & \colorbox{yh}{31.49} / \colorbox{gh}{0.96} \\
    \hline
    \end{tabular}}
    \caption{Comparison on our synthetic static testing data sorted by body-part. We compute average PSNR$\uparrow$ / SSIM$\uparrow$.}
    \label{tab:breakdown}
    \vspace{-3mm}
\end{table}
}

For video testing, in Tab.~\ref{tab:table2}, we evaluate our model on scenarios 1, 2, and 3 for both fidelity and temporal consistency. Other approaches face challenges in achieving both relit fidelity and temporal consistency at the same time as also shown in Fig.~\ref{fig:video_compare}. In contrast, our temporal module ensures our comprehensive relighting model to produce videos with strong temporal consistency. 
For more results on relighting results and comparisons including user study, please refer to the Supple.

\noindent\textbf{Ablation Study.} We conduct two ablation studies on our coarse-to-fine model using static human data and our temporal modules using dynamic human data. As shown in Tab.~\ref{tab:static_ablation}, we perform the ablation study on our coarse-to-fine approach: 1) \textit{Ours-diffusion}: Relighting only with pixel-aligned coarse shading without diffusion model, $\textit{i.e.}$, ${\mathbf{S}}_{\phi}\times\mathbf{I}$. 2) \textit{Ours end-to-end}: Relighting in an end-to-end manner by applying target lighting parameters $\boldsymbol\phi$ as a condition, trained with a diffusion model, without the coarse stage. 3) \textit{Ours-background}: Relighting without the control of background $\mathbf{B}$. 4) \textit{Ours+refine} (full): Applying our guided refinement to the relit image.
\begin{table}[tb]
    \centering
    \scalebox{0.98}{
    \begin{tabular}{
        p{2.8cm}<{\arraybackslash}|
        p{1.2cm}<{\centering\arraybackslash}|
        p{1.2cm}<{\centering\arraybackslash}|
        p{1.2cm}<{\centering\arraybackslash}} 
    \hline
    Method & $L_1$$\downarrow$ & PSNR$\uparrow$ & SSIM$\uparrow$ \\ 
    \hline
    Ours-diffusion & 0.05837 & 17.099 & 0.833\\
    Ours end-to-end  & 0.01432 & 26.418 & 0.918 \\
    Ours-background  & 0.01239 & 27.346 & 0.945 \\
    Ours & 0.01035 & 28.419 & 0.948\\
    Ours+refine & \cellcolor{yh}{0.01012} & \cellcolor{yh}{28.778} & \cellcolor{yh}{0.949} \\    
    \hline
    \end{tabular}}
    \caption{Ablation study on coarse-to-fine relighting models.}
    \vspace{-5mm}
    \label{tab:static_ablation}
\end{table}

\begin{table}[tb]
    \centering
    \scalebox{0.96}{
    \begin{tabular}{
        p{3.0cm}<{\arraybackslash}|
        p{1cm}<{\centering\arraybackslash}|
        p{1.3cm}<{\centering\arraybackslash}|
        p{1.3cm}<{\centering\arraybackslash}} 
    \hline
    Method & t$L_1$$\downarrow$ & tPSNR$\uparrow$ & tSSIM$\uparrow$ \\ 
    \hline
    Ours & 6.552 & 31.028 & 0.956\\
    Ours+temporal  & 5.638 & 32.266 & 0.957\\
    \hline
    Ours+temporal+blend  & \cellcolor{yh}{4.019} & \cellcolor{yh}{33.588} & \cellcolor{yh}{0.957} \\
    \hline
    \end{tabular}}
    \caption{Ablation study on our temporal modules evaluated on synthetic sequences: t$L1$ error ($\times 10^{-3}$)}
    \vspace{-6mm}
    \label{tab:video_ablation1}
\end{table}

Tab.~\ref{tab:static_ablation} shows the summary of the ablation study: Directly apply the coarse stage without a diffusion model introduces significant errors in the final relit result due to the detection noises. Instead of applying target lighting and end-to-end training with a diffusion model, our coarse-to-fine approach shows better performance, indicating that our coarse stage serves as a strong control prior. This control prior is both neat and effective for extending our model to diverse identities, various body parts. Based on the comparison with ``Ours-background'' we notice that encoding information from the background image aids in enhancing natural illumination during background harmonization. Lastly, leveraging a guided refinement enables the preservation of high-frequency information alongside robust generation capabilities.

For our temporal module, we study three ablation studies: 1) Ours: We eliminate all temporal consistency components, which is a single-frame-based generation method. 2) Ours+temporal: We only apply temporal module $\mathcal{E}_{m}$ without recurrent feature blending during the test-time. 3) Ours+temporal+blend: We perform our video relighting with temporal module $\mathcal{E}_{m}$ and recurrent feature blending. 

Tab.~\ref{tab:video_ablation1} summarizes the performance of each of our temporal modules. The temporal lighting module in ``Ours+temporal'' primarily enforces temporal coherence by imposing a temporal constraint on the lighting control between the current and previous frames during testing. Additionally, the recurrent blending feature further enhances temporal consistency by blending the lighting control feature between previous and current frames, thereby reinforcing the temporal context. 

\noindent\textbf{Limitation.} Our relighting diffusion model requires heavy computational time. Significant noise on the detection (\textit{e.g.}, mask and surface normal) affects the temporal coherence.

\section{Conclusion}%
\label{sec:Conclusion}
We introduce a method for Comprehensive Relighting that is generalizable and consistent for monocular human relighting and harmonization.
We address a core dataset challenge by utilizing a large and general image prior from a pre-trained diffusion model; and repurposing the model specialized for temporally consistent image relighting.
For coherent control of the lighting, we introduce a coarse-to-fine relighting framework; and combine it with an external temporal lighting module that learns many real videos.
Our guided refinement network enhances the visual to preserve the fine details of an original image. 
In the experiments, our method outperforms other image-based relighting and harmonization models in terms of quality and temporal coherence.
\label{sec:conclusion}

\section{Acknowledgement}
We sincerely thank Mengwei Ren for the insightful discussions regarding the framework design, and we are grateful to Jianming Zhang for kindly providing the human normal map estimator.


{\small
\bibliographystyle{ieeenat_fullname}
\bibliography{references}
}

\ifarxiv \clearpage \appendix 

\begin{appendices}
In this document, we provide more details for the method, experiments, dataset, and more qualitative results, as an extension of Sec. 3 and Sec. 4 in the main paper. Please also refer to the video demo for dynamic relighting results, comparison, ablation study, and more results. 
\footnotetext[1]{$^\dagger$This work is partially done during an internship at Adobe Research.}

\section{Method and Experiment Details}

We demonstrate that during training, instead of directly using albedo and shading maps, we train with relit images using different lighting augmentations. By leveraging a conditional diffusion model, our approach can implicitly disentangle lighting and appearance from the input image, learning to generate relit images and bypassing the need for a preprocessed de-lighting process.

\subsection{Relighting and Harmonization Diffusion Network (Sec. 3.2)}

\begin{figure}[H]
\centering
\includegraphics[width=0.475\textwidth]{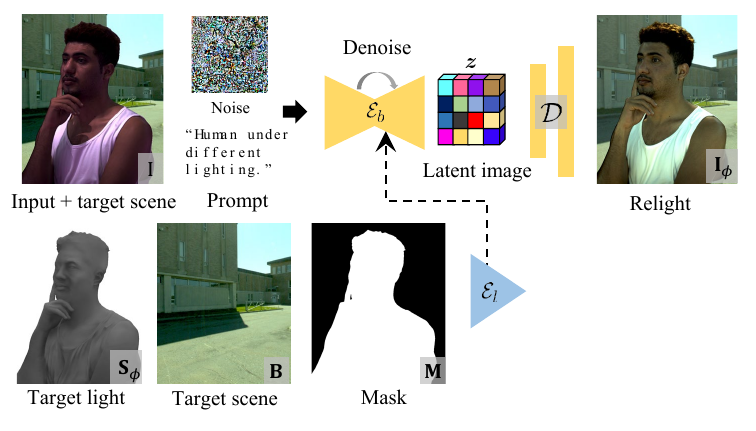}
\caption{Relighting and Harmonization diffusion model training and denoising pipeline.}
\label{fig:diffusion_pipeline}
\end{figure}

As shown in Fig.~\ref{fig:diffusion_pipeline}, which includes the diffusion model training process and denoising (sampling) process for our fine-grained relighting. During the training process, we follow the same Stable Diffusion architecture as~\cite{brooks2023instructpix2pix}, and both Lighting ControlNet and Motion ControlNet architecture are followed by~\cite{zhang2023adding}. Stable Diffusion model adopts a U-Net~\cite{ronneberger2015u} architecture comprising an encoder, a middle block, and a skip-connected decoder. Each of the encoder and decoder consists of 12 blocks, totaling 25 blocks within the complete model, and each primary block integrates 4 ResNet layers and 2 Vision Transformers (ViTs) with cross-attention and self-attention mechanisms. The ControlNet architecture is applied at each encoder level of the U-Net, featuring a trainable copy of 12 encoding blocks and 1 middle block from the Stable Diffusion model. These 12 encoding blocks includes: 64 $\times$ 64, 32 $\times$ 32, 16 $\times$ 16, 8 $\times$ 8, with each resolution replicated 3 times. The resulting outputs are merged with the 12 skip connections and the single middle block within the U-Net structure. We fine-tune both ControlNet and Stable diffusion module on our relighting dataset.

\subsection{Training Dataset (Sec. 4)}
In Fig.~\ref{dataset}, we visualize the samples of our training dataset. We use two kinds of dataset. One is from the data captured from LightStage where the background images are rendered from a HDR environment map. The ground truth shading, albedo, relighted image, and background captured from a small number of viewpoints (\textit{e.g.}, 6 views) are available. The other one is from the data rendered from a synthetic human model. We render the image of many 3D human models from many views (e.g., 16 views) under different lighting conditions defined by an environment map. We obtain the approximated spherical harmonics coefficients from the environment maps as ground-truth lighting parameters. The ground truths for the mask, albedo, background, and relit images also exist.
\begin{figure*}[tb] \centering
    \includegraphics[width=\textwidth]{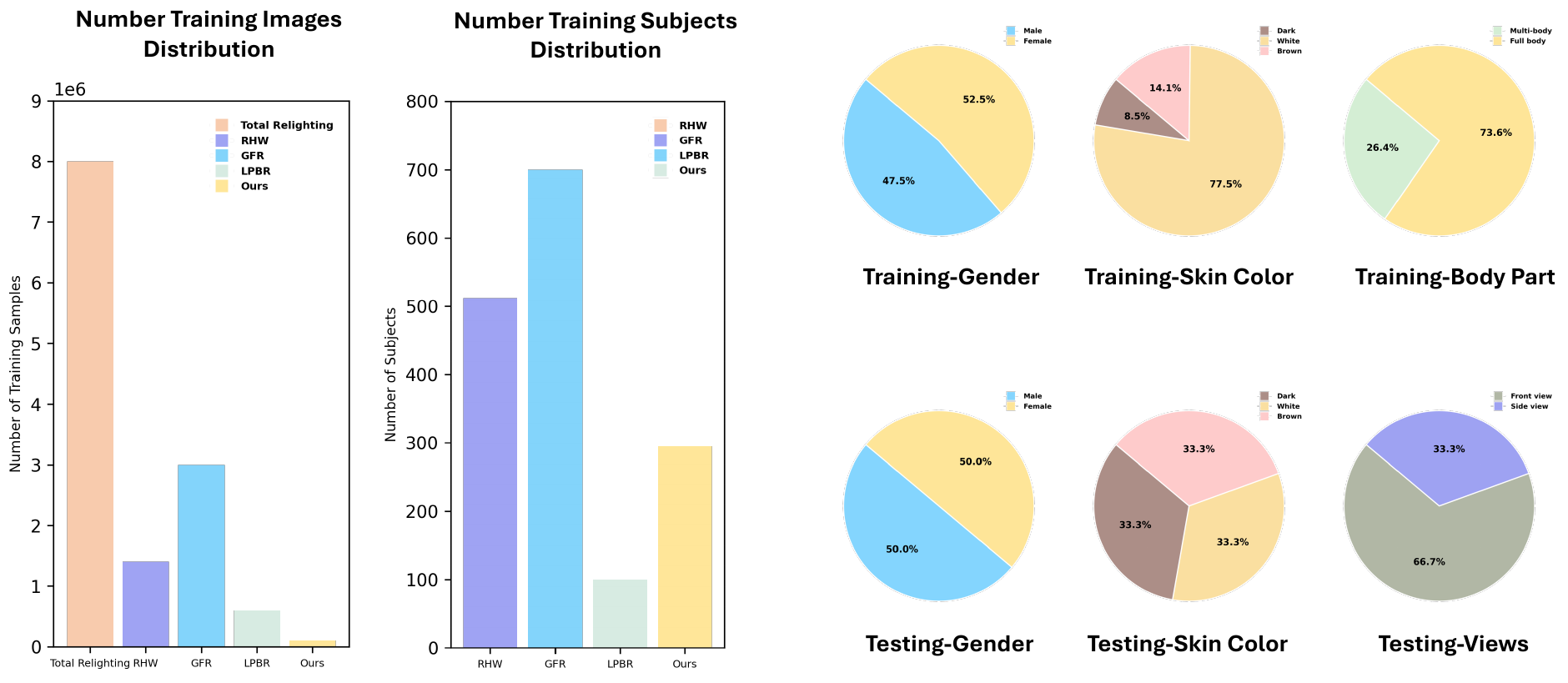}
    \caption{Left side: Training data scale comparisons; Right side: Breakdown of our training and evaluation dataset information.} \label{fig:dataset}
\end{figure*}
We kindly note that our training data is relatively smaller compared to other image-based relighting methods as summarized in Fig.~\ref{fig:dataset}. For instance, Total Relighting~\cite{pandey2021total} captures data from 70 diverse subjects. Through extensive lighting augmentation, the dataset expands to include approximately 8 million OLAT training examples; GFR~\cite{ji2022geometry} needs 700 subjects and 4,600 HDR maps for training; and LPBR~\cite{ren2023relightful} is trained on 100 subjects with OLAT and 2,908 HDR maps, resulting in 600K training samples. Our training data is composed of 100K samples where the detailed data analysis can be found in Fig.~\ref{fig:dataset}.
We categorize our training data based on gender, skin tone, and body coverage (half-body and full-body). Each subject is captured from 32 viewpoints under varying lighting conditions.
\subsection{Add-on Temporal Motion Module Network (Sec. 3.3)}
\begin{algorithm}[H]
 \caption{Unsupervised Cycle-Training Motion Modeling for Temporal Consistency}
  \label{alg:cycle-train}
  \begin{algorithmic}[1]
    \STATE \textbf{Require}: Video frames $\mathbf{I}$; decoder $\mathcal{D}_{*}$
    \STATE \textbf{Require}: Relit frames $\mathbf{I}_{\phi} \gets (\mathcal{D}_{*} \circ \mathcal{E}_{b})$
    \STATE \textbf{Initialize}: Motion encoder $\mathcal{E}_{\rm m}$; train step function $\mathbf{T}$
    \STATE Converged $\gets$ \textbf{False}
    \STATE \textbf{While} not Converged \textbf{do}
    \STATE \hspace{5mm} $\mathbf{I}^{t}_{\phi} \gets \mathcal{D}^{*}(\mathcal{E}_{\rm b}^{*}(\mathbf{I}^{t}, \mathcal{E}_{\rm l}^{*}(\{\mathbf{S}_{\phi}^{t}, \mathbf{B}^{t}\}, \mathbf{M}^{t})))$
    \STATE \hspace{5mm} $\tilde{\mathbf{I}}^{t}_{t-1} \gets \mathcal{D}^{*}(\mathcal{E}^{*}_{\rm b}(\mathbf{I}^{t}_{\phi}, \mathcal{E}_{\rm m}(\mathbf{I}^{t-1},\mathbf{M}^{t-1})))$
    \STATE \hspace{5mm} Converged $\gets \mathbf{T}(\tilde{\mathbf{I}}^{t}_{t-1}, \mathbf{I}^{t})$
    \STATE \textbf{end while}
  \end{algorithmic}
\end{algorithm}

We present the cycle-training algorithm for our temporal lighting module in Alg.\ref{alg:cycle-train}, which serves as an additional explanation for Sec. 3.3. Based on the hypothesis: original video sequence inherently contains temporal lighting properties, which can be modeled by a temporal module, conditioned on the relit version. We train an add-on temporal module in an unsupervised way. Before the training process, we require relit video frames, $\mathbf{I}^{t}\rightarrow\mathbf{I}^{t}_{\phi}$. To generate the relit frame we process forward image relighting: $\mathbf{I}^{t}_{\phi}\gets\mathcal{D}^{*}(\mathcal{E}_{\rm b}^{*}(\mathbf{I}^{t};\mathcal{E}_{\rm l}^{*}(\{\mathbf{S}_{\phi}^{t}, \mathbf{B}^{t}\}; \mathbf{I}^{t},\mathbf{M}^{t})))$. 
During each training iteration, as indicated in: $\tilde{\mathbf{I}}^{t}_{t-1}\gets\mathcal{D}^{*}(\mathcal{E}^{*}_{\rm b}(\mathbf{I}^{t}_{\phi};\mathcal{E}_{\rm m}(\mathbf{I}^{t-1},\mathbf{M}^{t-1})))$, 
we condition on the current relit frame and revert the lighting of the previous frame in the original video back to match that of the original frame. 

\begin{figure*}[t]
\centering
\includegraphics[width=0.9\textwidth]{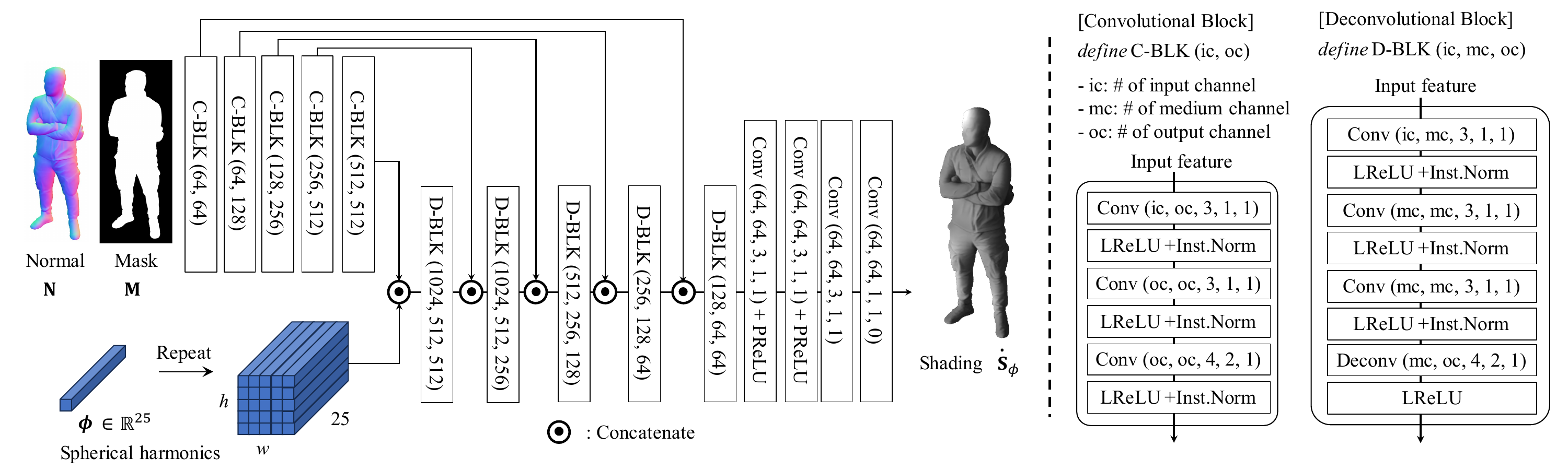}
\caption{Left: Our shading estimation network, Right: Convolutional and deconvolutional blocks.} 
\label{fig:shading_net}
\end{figure*}

\begin{figure*}[t]
\vspace{-3mm}
\centering
\includegraphics[width=0.95\textwidth]{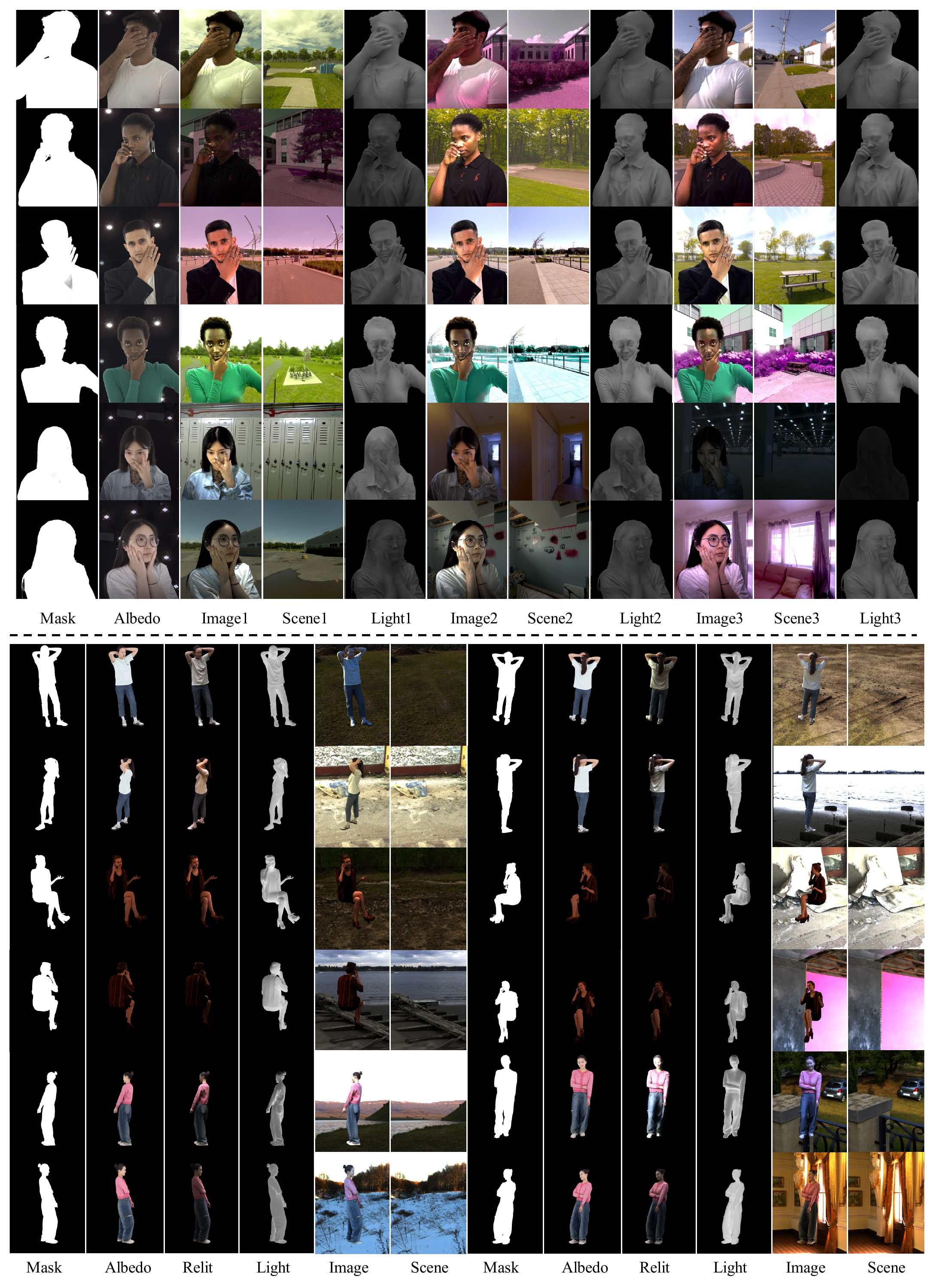}
\vspace{-3mm}
\caption{Training samples of the relighting data with half-body portraits (up) and simulation data with full-body images (bottom) .} 
\label{dataset}
\vspace{-3mm}
\end{figure*}

\noindent\textbf{Implementation details.} We train our model on 8 A100 GPUs with a total batch size of 32 (4 batches per GPU) and a learning rate of $2 \times 10^{-6}$. In the training phase for Lighting ControlNet, we initialize the Stable diffusion base model using the pre-trained weights from Instruct-Pix2Pix~\cite{brooks2023instructpix2pix}, and copy the encoder block weights to serve as the initial weights for the Lighting ControlNet part. Subsequently, we fine-tune both ControlNet and Stable Diffusion module on our relighting dataset

The training of our Motion ControlNet module occurs subsequent to the lighting control training process. During the training phase for motion control, we freeze the weights of the Stable Diffusion base model. Then, we initialize the weights of the Motion ControlNet by copying the encoder block weights from the previously trained lighting Stable Diffusion. Subsequently, we exclusively fine-tune the Motion ControlNet.

During the inference process, we adopt random noise with a resolution of 4 $\times$ 96 $\times$ 96 as the initial input to generate the final relit image with a resolution of 768 $\times$ 768, and for video testing, we apply the same noise across frame. We apply DDIM~\cite{song2020denoising} sampler with a timestep of 50 to generate the final relit image. To utilize frame-by-frame inference with recurrent blending, we extract control features from the 12 encoding blocks of the ControlNet at corresponding resolutions. Subsequently, we perform weighted blending between control feature of previous and current frames.

\subsection{Pixel-Aligned Neural Shading (Sec. 3.2)}
While coarse shading ${\mathbf{S}}_{\phi}$ can be directly computed from Spherical harmonics (SH) lighting parameters, we experimentally found that using ${\mathbf{S}}_{\phi}$ obtained from a neural network can improve human relighting and harmonization. Specifically, low-order SH models tend to smooth out fine details, resulting in overly diffuse shading. In contrast, a neural network can recover high-frequency shading variations, enhancing realism by capturing subtle lighting effects. Moreover, the learned shading function improves robustness to normal map inaccuracies, reducing artifacts and better preserving surface details. In this section, we introduce an alternative way of having a coarse shading using a neural network. To this end, we introduce a pixel-aligned lighting estimation function $f$ in Eq. 2 using a conditional Unet framework.

It takes as inputs surface normal map $\mathbf{N}$ and target lighting parameters $\boldsymbol{\phi}$ as conditions, and estimates the shading ${\mathbf{S}}_{\phi}$ at each pixel lit by the target lighting.   
$\mathbf{N}$ is detected from the input image $\mathbf{I}$ using the internal normal detector which is composed of Unet architecture with pyramid vision transformer~\cite{wang2022pvt}.
It learns many mixtures of ground-truth data similar to~\cite{ranftl2020towards}, and thus, applicable to general scenes and objects.  
Note that, since $f$ does not take any visual data as inputs, it does not introduce visual domain gaps.
We train the $f(\cdot)$ by comparing the input image and its reconstruction from the estimated shading:
%
\begin{equation}
\mathcal{L}_\mathrm{recon}=\sum_{i} \|\mathbf{I}_{\rm recon}-\mathbf{I}\|^{\rm 2}_{\rm 2}=\sum_{i}\|{\mathbf{S}}_{\phi}\odot\mathbf{A}_{\rm GT}-\mathbf{I}\|^{\rm 2}_{\rm 2}\nonumber
\label{ep5}
\end{equation}
where $\mathbf{I}_{\rm recon}$ is the reconstructed image based on the multiplication of $\dot{\mathbf{S}}_{\phi}$ with the ground-truth albedo $\mathbf{A}_{\rm GT}\in\mathbb{R}^{w\times h\times 3}$. Since we supervise the shading estimation network in the image space, we can utilize other advanced image-based supervision signals that can capture the physical plausibility of the local and global shading as follows:
\begin{equation}
L_\mathrm{\rm shade}=\mathcal{L}_\mathrm{recon}+\lambda_\mathrm{v}\mathcal{L}_\mathrm{vgg}+\lambda_\mathrm{c}\mathcal{L}_\mathrm{cGAN},
\label{eq5}
\end{equation}
where $L_\mathrm{shade}$ is the entire objective, and $\lambda$ controls the weight of each loss function. $\mathcal{L}_\mathrm{vgg}$ is designed to penalize the difference between the reconstructed image $\mathbf{I}_{\rm recon}$ and the input $\mathbf{I}$ in the deep feature space~\cite{johnson2016perceptual}. $\mathcal{L}_\mathrm{cGAN}$ is the conditional adversarial loss~\cite{isola2017image} to evaluate the plausibility of the reconstructed shading with respect to the geometric structure where we use $\{\mathbf{N}, \ \mathbf{I}\}$ as real and $\{\mathbf{N}, \ \mathbf{I}_{\rm recon}\}$ as fake conditions to the patch discriminator~\cite{isola2017image}.

\begin{figure*}[tb] \centering
    \includegraphics[width=\textwidth]{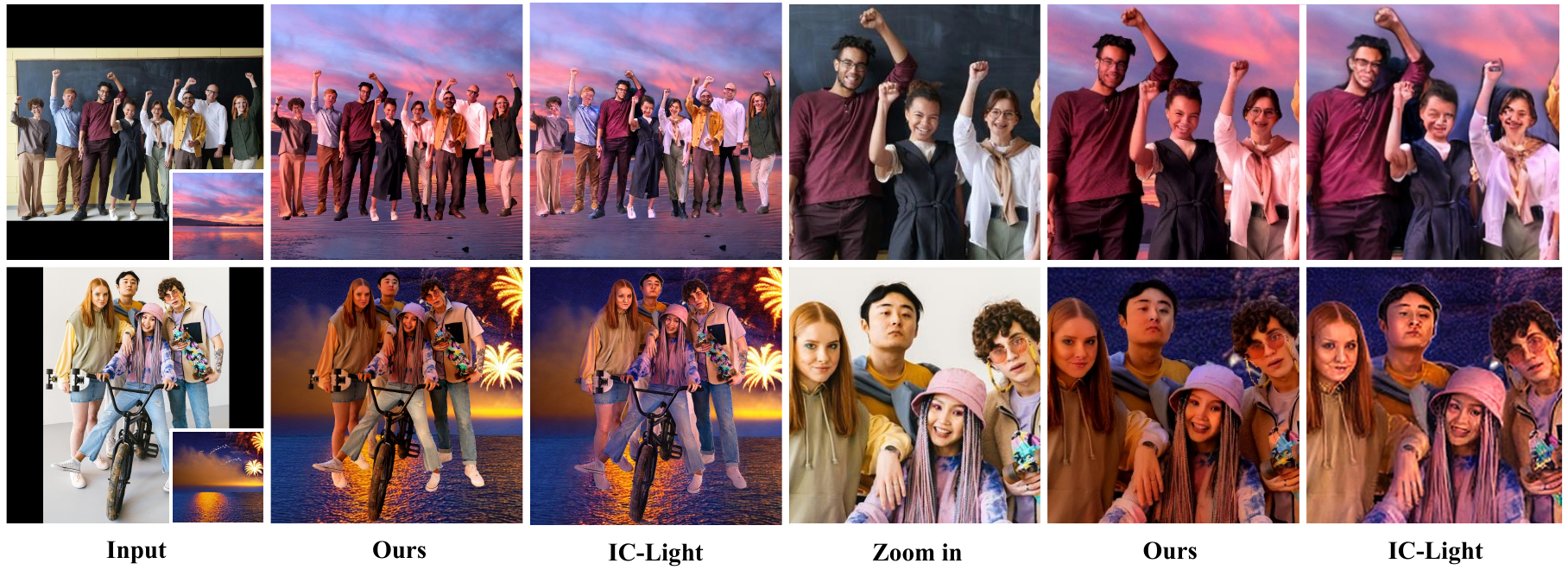}
    \caption{Comparison with harmonization methods (IC-Light). Left side is multi-person testing, right side is zoom in result.} \label{fig:img3}
\end{figure*}

\noindent\textbf{Coarse Shading Estimation Network}. 
In Fig.~\ref{fig:coarse}, we show the general training pipeline for coarse lighting estimation network. Fig.~\ref{fig:shading_net} describes the structure of our coarse shading estimation network. It takes as inputs the surface normal, foreground mask, and lighting parameters (\textit{i.e.}, Spherical harmonics); and generates the shading map. An encoder regresses the surface normal and mask to the latent space. In this latent space, the lighting parameters are conditioned where the vector parameters are copied along the spatial direction to fit the same latent space as the one from the encoder. A decoder decodes them to generate a shading map.
\begin{figure}[H]
\centering
\includegraphics[width=0.475\textwidth]{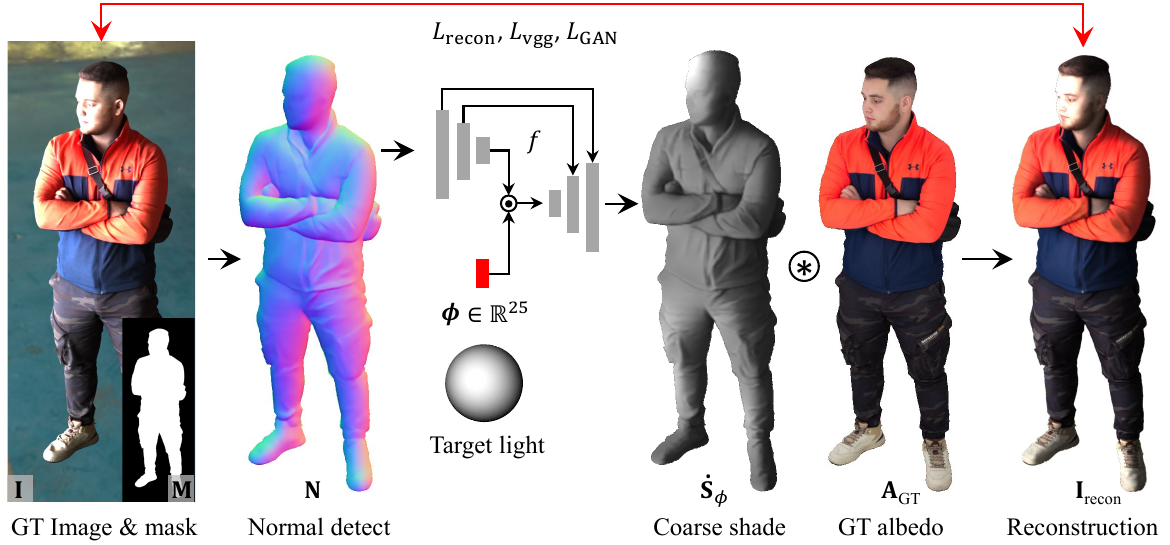}
\caption{Training pipeline for coarse lighting estimation network.}
\label{fig:coarse} 
\end{figure}

\section{Qualitative Results}
\vspace{-2mm}
\subsection{Comparison with other baselines (Sec. 4)} 
We present the qualitative results of static image testing on our synthetic dataset, compared with other baseline methods: DPR~\cite{DPR}, GFR~\cite{ji2022geometry} and RHW~\cite{tajima2021relighting} in Fig.~\ref{fig:supple_static}. In our evaluation, we perform full-body and multi-person tests on our synthetic testing dataset, integrating background images alongside Spherical harmonics for lighting control. We calculate the average error on the entire testing dataset for a comprehensive and generalizable relighting evaluation. From visual quantitative results, our model shows more realistic relighting results compared to other human relighting models. This demonstrates our model's robust performance across diverse body part tests, indicating higher generalizability. 

For evaluation, we validate our model along with other baselines based on the divided categories: gender, and skin color. We present the numerical evaluation in Tab.~\ref{tab:compare2} and~\ref{tab:compare3}. From the qualitative results, our method consistently outperforms in all categories.

We further highlight that while all those methods are limited to working on a specific body part (e.g., face or portrait), our method works on general cases including the scene with face, portrait, full body, and multi-person.

\begin{figure*}[hbt!]
\centering
\includegraphics[width=1\linewidth]{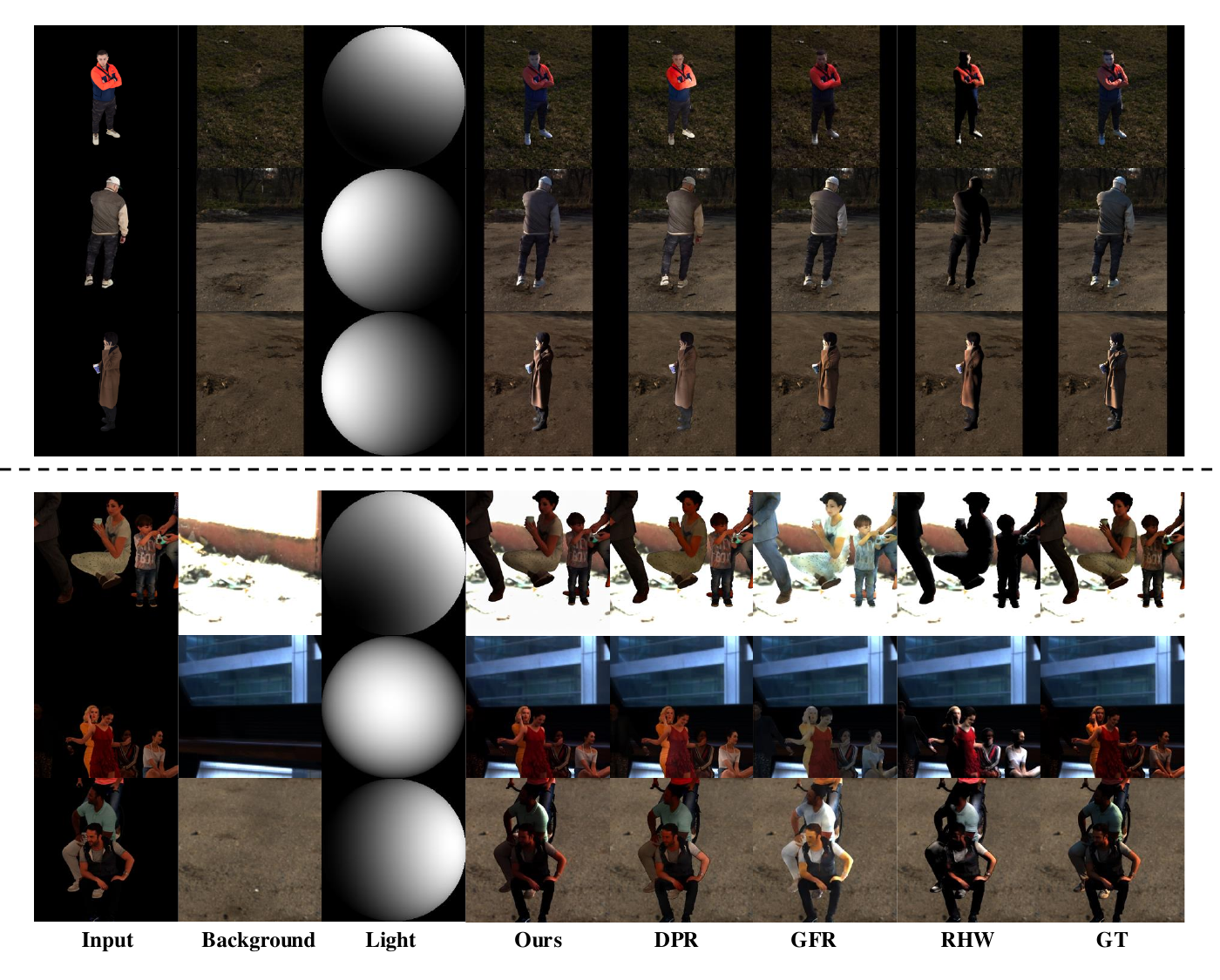}
\vspace{-1mm}
\caption{Qualitative comparisons conducted on synthetic data. From top to bottom: full-body testing, multi-person testing. The ground truth data is displayed in the last column.}
\label{fig:supple_static}
\vspace{-2mm}
\end{figure*}

We present real data comparison results on the LightStage dataset in Fig.~\ref{fig:compare_img2} and comparisons on in-the-wild images in Fig.~\ref{fig:compare_img1}. Since current state-of-the-art (SOTA) baselines are not designed for comprehensive relighting, their performance varies across different scenarios. In Fig.~\ref{fig:compare_img1}, while DPR performs well for face relighting, its quality significantly deteriorates in half-body scenarios, exhibiting strong artifacts due to domain gaps. Notably, our framework is the first to achieve comprehensive relighting, effectively handling arbitrary body parts, including portraits, half-body, full-body, and multi-body scenarios.

In Fig.~\ref{fig:real_static1} and Fig.~\ref{fig:real_static2}, we present static real image relighting and harmonization comparison results. For harmonization, we use the most recent work, LPBR~\cite{ren2023relightful}, as one of the baselines: (1) DPR and RHW are only applicable to image relighting with Spherical harmonics for lighting control. For a fair comparison, we tested image relighting with DPR, RHW, and GFR in Fig.~\ref{fig:real_static1}, using a black background and target lighting parameters. We applied different lighting conditions to various identities, including half-body and full-body images. Although these methods can achieve human relighting, their limited generalizability results in less fidelity during comprehensive testing. (2) Both LPBR and GFR can perform harmonization. We retrained the GFR model with our settings, enabling it to achieve both harmonization and relighting, as shown in Fig.~\ref{fig:real_static2}. The higher generative prior of LPBR, which also uses a diffusion model, results in noticeable distortions on the human face. Although GFR can achieve both harmonization and relighting, it exhibits obvious color noise.

In Fig.~\ref{fig:img3}, we present a new comparison with IC-Light~\cite{zhang2025scaling}, which is the current state-of-the-art for light-aware background harmonization. Both IC-Light and our model are stable diffusion relighting models. IC-Light can generate relit images with text prompts or background harmonization. In the visual results, our harmonization seamlessly blends with the target background while preserving the original identity. While IC-Light also achieves high-quality background harmonization, however, it exhibits greater identity distortion at the same image resolution, particularly in full-body and multi-person scenarios. In Fig.~\ref{fig:img1}, third graph, we show the user preference comparison among our method, LPBR, and IC-Light. Most users selected our method as the best result for all questions.

For video relighting comparison, we present qualitative results in Fig.~\ref{fig:supple_video}, in the main paper. We show frames relit by our model tested on the synthetic video testing data. The first row shows the composite input (albedo foreground and background). In the second row, we show the ground truth shading, and the third row displays the ground truth relit image. The following rows show our relit frames, followed by those from GFR, RHW, LPBR, and DPR. For real video comparison, please refer to the supplementary demo video.

{\renewcommand{\tabcolsep}{6.5pt}
\begin{table}
    \centering
    \scalebox{1.0}{
    \begin{tabular}{|l||c|c|c|c|}
    \hline
    \textbf{Method} & \textbf{SH} & \textbf{Bg} & \textbf{Male} & \textbf{Female} \\
    \hline
    RHW & \cmark & \xmark & 28.89 / 0.950  & 26.58 / 0.939  \\
    DPR & \cmark & \xmark & 27.63 / 0.972  & 27.62 / 0.944  \\
    GFR & \cmark & \cmark & 29.32 / 0.926  & 29.71 / 0.973  \\
    Ours & \cmark & \cmark & \textbf{31.12} / \textbf{0.970}  & \textbf{30.50} / \textbf{0.964} \\
    \hline
    \end{tabular}}
    \caption{Comparison of baseline methods on our full-body synthetic static data, categorized by gender: (PSNR$\uparrow$ / SSIM$\uparrow$).}
    \vspace{-2mm}
    \label{tab:compare2}
\end{table}
}

{\renewcommand{\tabcolsep}{5.pt}
\begin{table}
    \centering
    \scalebox{0.98}{
    \begin{tabular}{|l||c|c|c|}
    \hline
    \textbf{Method} & \textbf{White} & \textbf{Brown} & \textbf{Dark} \\
    \hline
    RHW & 28.15 / 0.946 & 27.37 / 0.944 & 27.68 / 0.943 \\
    DPR & 27.44 / 0.956 & 27.70 / 0.962 & 27.73 / 0.956 \\
    GFR & 29.94 / 0.936 & 29.41 / 0.934 & 29.10 / \textbf{0.978} \\
    Ours & \textbf{31.53 / 0.985} & \textbf{31.77 / 0.976} & \textbf{29.13} / 0.940 \\
    \hline
    \end{tabular}}
    \caption{Comparison of baseline methods on our full-body synthetic static data, categorized by skin color: (PSNR$\uparrow$ / SSIM$\uparrow$).}
    \vspace{-2mm}
    \label{tab:compare3}
\end{table}
}

\begin{figure*}[tb] \centering
    \includegraphics[width=\textwidth]{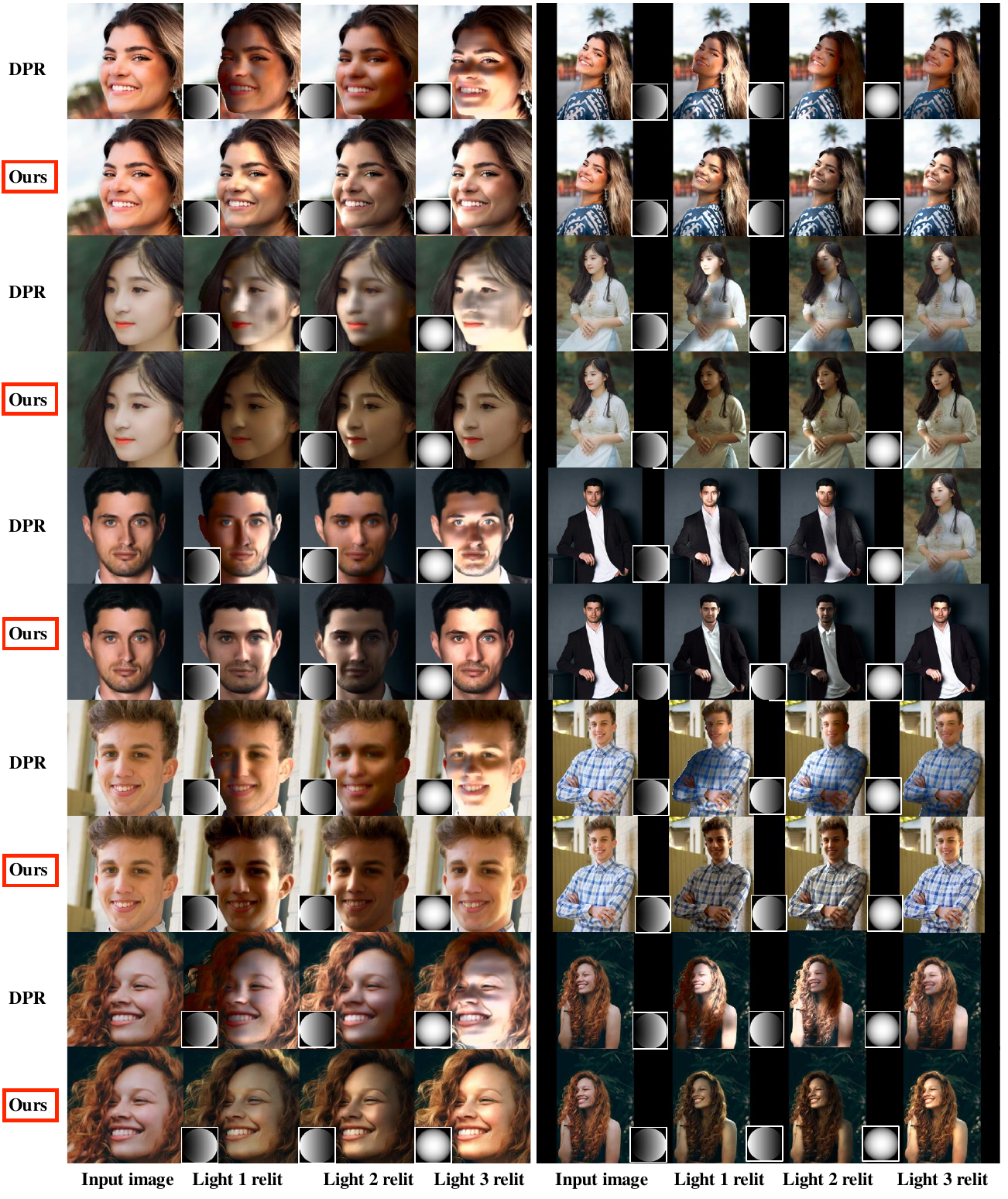}
    \caption{Comparison with DPR on face and half-body relighting on Pexels~\cite{Pexels} real images.} \label{fig:compare_img1}
    \vspace{-1em}
\end{figure*}

\begin{figure*}[tb] \centering
    \includegraphics[width=\textwidth]{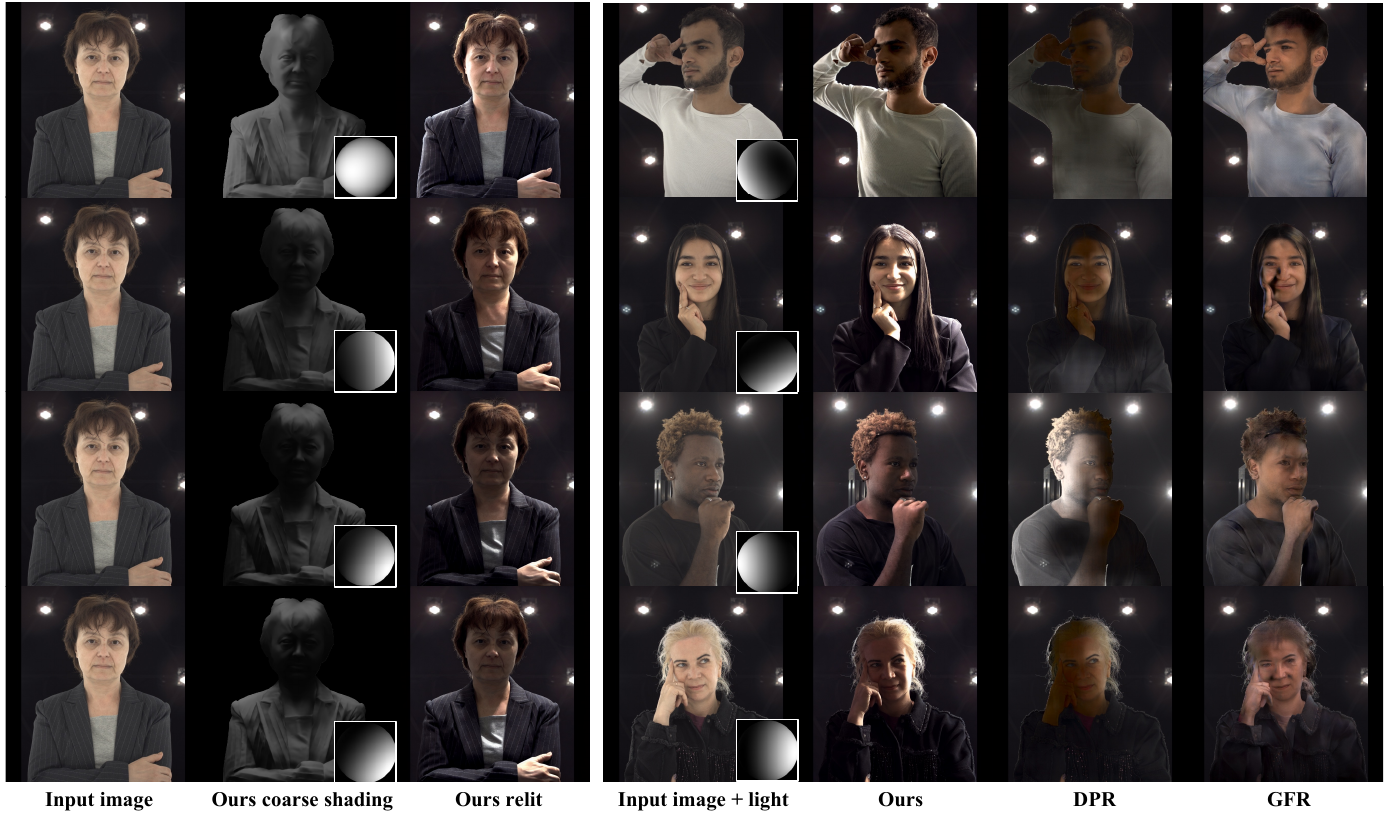}
    \caption{Our LigtStage data testing (Left) and comparison with other relighting baselines (Right).} \label{fig:compare_img2}
    \vspace{-1em}
\end{figure*}

\begin{figure*}[tb] \centering
    \includegraphics[width=\textwidth]{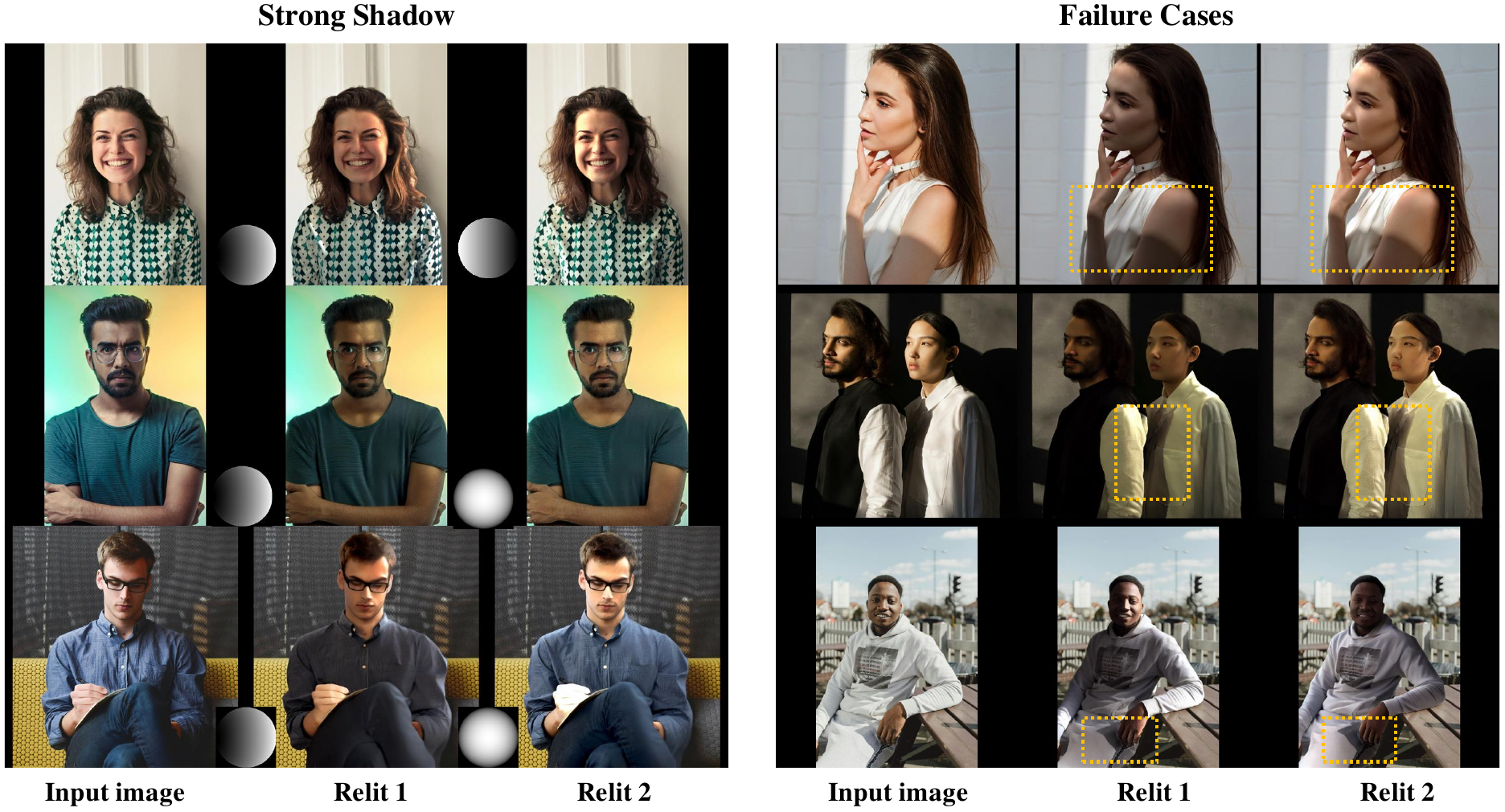}
    \caption{Strong shadow testing results (left) and failure cases (right) on real images from Pexels~\cite{Pexels}.} \label{fig:img2}
    \vspace{-1em}
\end{figure*}


\begin{figure*}[t]
\centering
\includegraphics[width=0.82\textwidth]{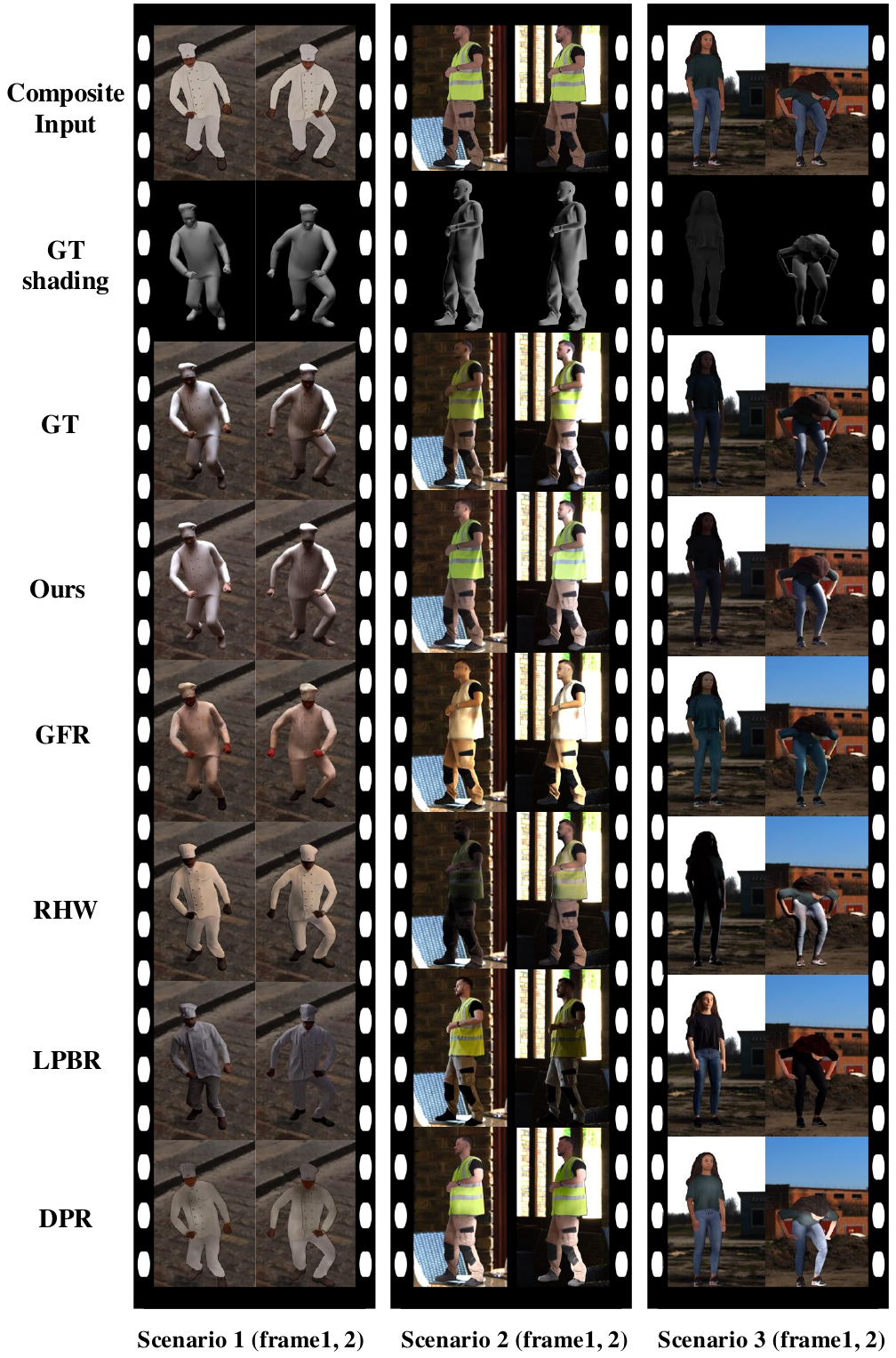}
\caption{Video relighting comparison results on synthetic testing data: from left to right, we show comparison results for Scenario 1, 2, 3. From top to bottom, the first row shows the composite input (foreground human albedo composited with background image), the second row shows the ground truth (GT) shading, and the third row shows the GT image.}
\label{fig:supple_video}
\end{figure*}

\begin{figure*}[t]
\centering
\includegraphics[width=0.78\textwidth]{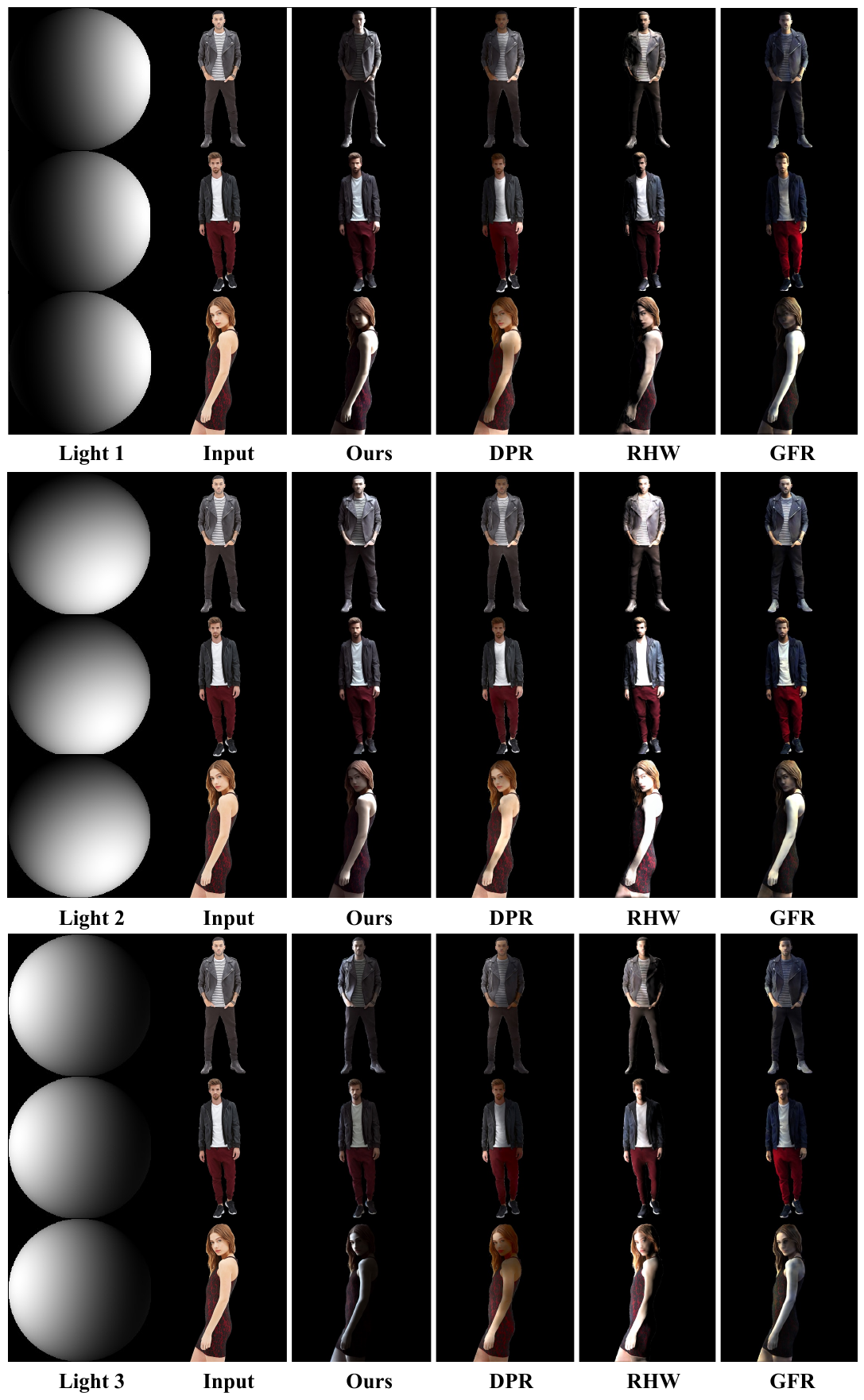}
\caption
{Real image comparisons with other human relighting approaches on the DeepFashion dataset~\cite{liu2016deepfashion}. We test on different identities and body parts (full body, half body). Our model shows consistent and feasible relighting with varying target lighting parameters (Spherical harmonics).}
\label{fig:real_static1}
\end{figure*}

\begin{figure*}[t]
\centering
\includegraphics[width=0.84\textwidth]{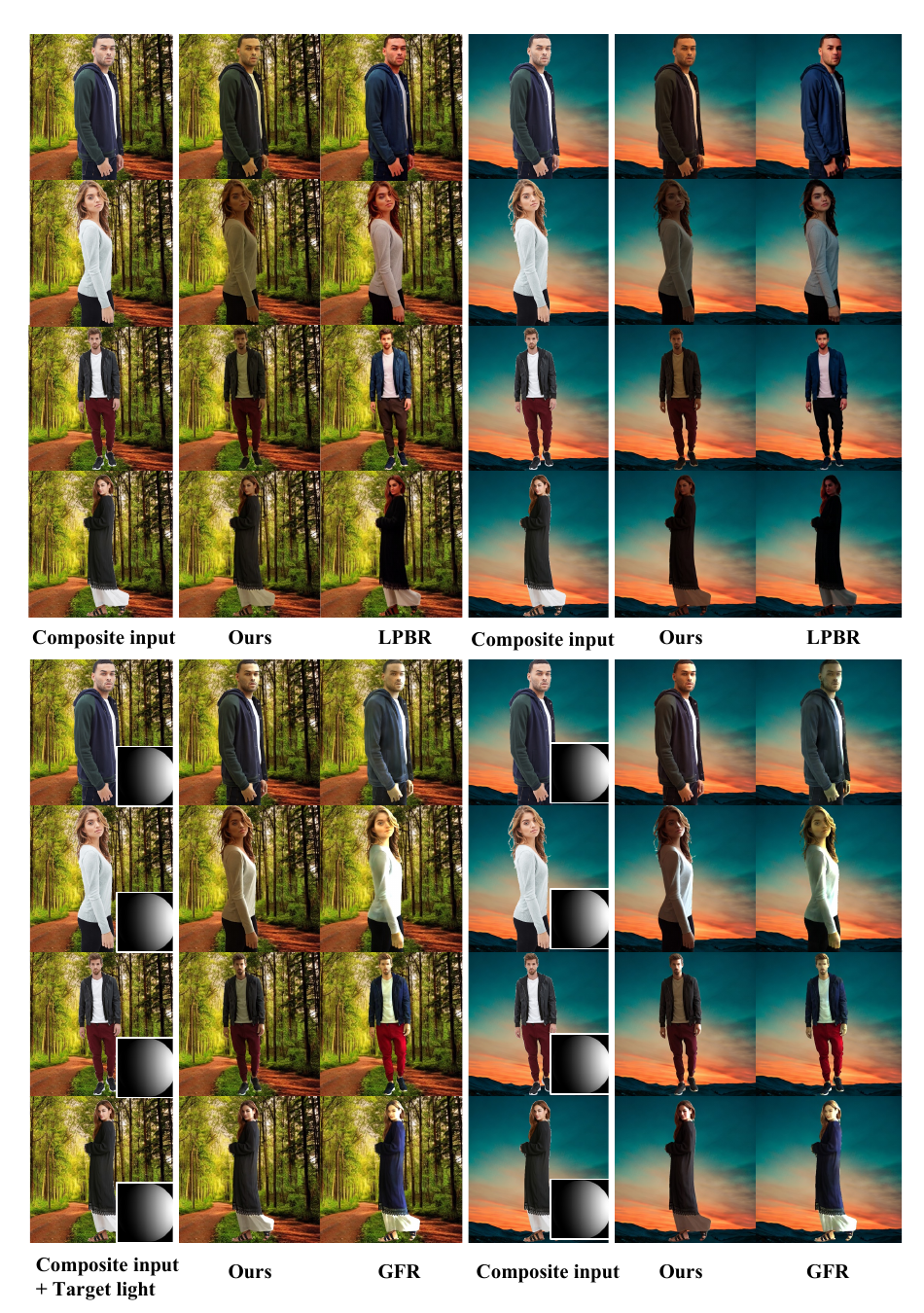}
\caption{We present real image comparisons with the harmonization method. Given a composite input image, our model can achieve effective harmonization. When provided with target lighting parameters (Spherical harmonics), our model can achieve both background harmonization and relighting. The top section displays the outputs of our background harmonization method compared to the results from \cite{ren2023relightful}. The lower section presents harmonization and relighting comparisons with \cite{ji2022geometry}. Due to the higher generative prior of LPBR, noticeable distortions are present on the human face. Although GFR can achieve both harmonization and relighting, it exhibits obvious color noise.}
\label{fig:real_static2}
\end{figure*}

\subsection{More qualitative results} We present additional qualitative results on the DeepFashion dataset~\cite{liu2016deepfashion}, as shown in Fig.~\ref{fig:ours_lighting1}. Given an input image (left side) and target lighting parameters, our model achieves the relighting results (second column). By changing the background image, our model can achieve both background harmonization and relighting, as demonstrated in columns 3 through 7.

Our model can achieve realistic relighting effects given a target lighting, as well as background harmonization and a combination of both. It effectively handles diverse subjects with varying identities and poses, including both half-body and full-body representations, demonstrating higher generalizability. 

\subsection{Performance and rendering time}
For the generation of the 768x768 pixel resolution image with stable quality, 50 diffusion timesteps are required, leading to around 10 seconds. For video sequences with relighting using a motion module, each frame takes approximately 25 seconds on an A100 GPU. In theory, there is no limit in the number of frames that our model can handle, the video rendering time is highly proportional to the number of frames, requiring around 2 hours for a video clip with 300 frames (768x768).

\subsection{User study}
\begin{figure}[H] 
\centering
    \includegraphics[width=0.475\textwidth]{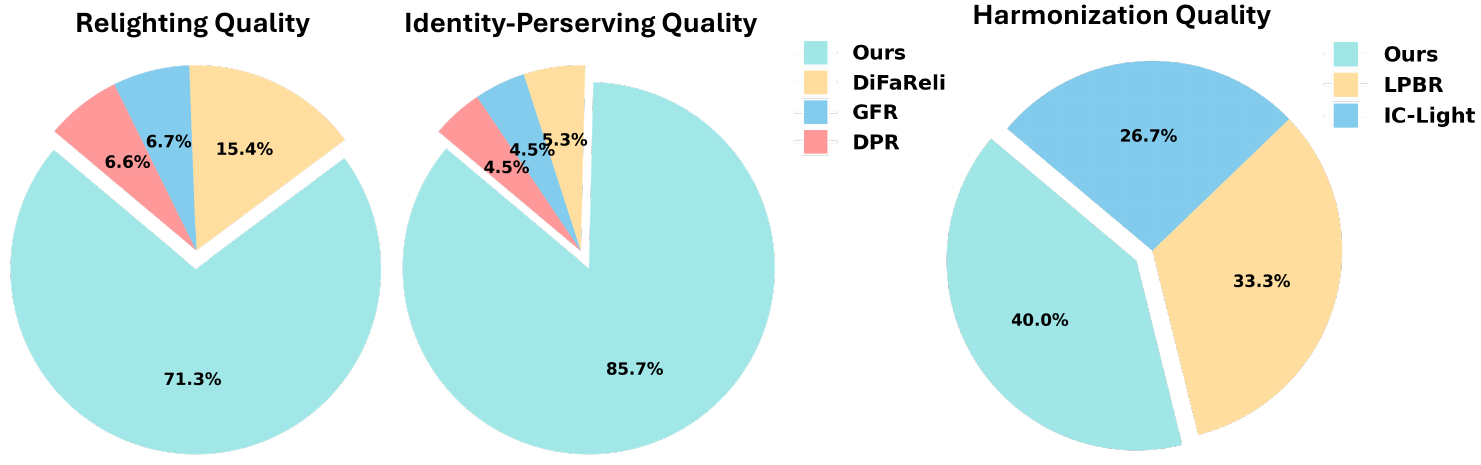}
    \caption{User study results: Preferences between our model and other relighting and harmonization models, including our general object testing. } \label{fig:img1}
    \vspace{-1em}
\end{figure}
We performed a user study as shown in Fig.~\ref{fig:img1}. For the relighting model, we used three state-of-the-art methods: DiFaReli~\cite{ponglertnapakorn2023difareli}, GFR~\cite{ji2022geometry}, and DPR~\cite{zhou2019deep}. For the harmonization model, we chose LPBR~\cite{ren2023relightful}. Users participated in answering three questions:

\begin{itemize}
    \item \textbf{Q1:} Which result most effectively achieves the relighting?
    \item \textbf{Q2:} Which result most effectively preserves the person’s identity (e.g., details and skin)?
    \item \textbf{Q3:} Which result best harmonizes with background scenes?
\end{itemize}

We summarized the percentage of user preferences and plotted the pie graph as shown in Fig.~\ref{fig:img1}. Overall, users selected our method as the best result for all questions, implying that our method is perceptually effective in achieving reasonable relighting quality, preserving identity, and harmonizing with the background.

\section{Limitation and future work } 
\noindent In Fig.~\ref{fig:img2}, we demonstrate some relighting results of the person under shadow and highlights. While our method can suppress shadows from self-occlusion during relighting, we acknowledge that our model shows some weaknesses with strong shadows, especially on human clothes (failure cases in Fig.~\ref{fig:img2}, right side). In fact, these strong shadows can be further suppressed by existing shadow removal models such as~\cite{yoon2024generative, futschik2023controllable, weir2022deep}. Additionally, incorporating various training data augmentations for hard shadows can be explored as future work to further enhance relighting quality.
Our relighting diffusion model requires significant computational time. Recent advancements in diffusion models, such as the One-Step Diffusion Model~\cite{yin2024one}, may further enhance inference efficiency.
Significant noise on the detection (\textit{e.g.}, mask and surface normal) affects the temporal coherence, and we admit that our results still have residual flickering. Nevertheless, our approach surpasses other relighting methods in video quality across diverse domains. We believe that advancing video prior models and expanding video datasets will further enhance temporal coherence, which we plan to explore in future work.
Our task primarily focuses on human relighting, which limits the model's ability to accurately handle materials associated with general objects such as cars, glass, and metallic surfaces. We acknowledge this limitation and plan to explore this aspect in future work.

\section{Broader Impact} 
\noindent As a positive impact, this work can be a useful tool for enhancing the lighting condition of the picture with humans, which can be useful for contents creation in social media. As a negative impact, similar to image synthesis, this work can synthesize human appearance under different lighting that may be used to fabricate fake videos and news.


\begin{figure*}[t]
\centering
\includegraphics[width=0.88\textwidth]{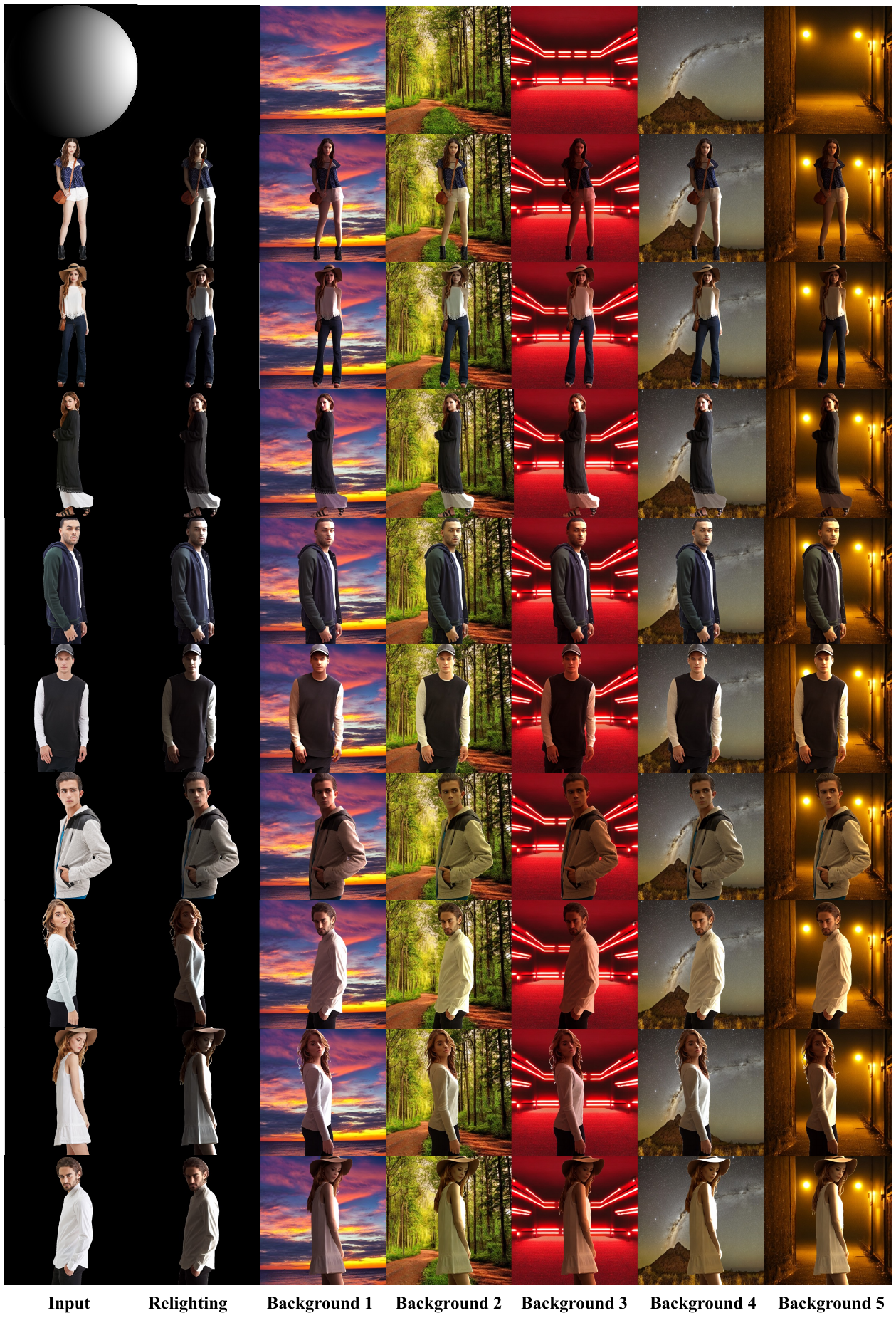}
\caption{Our model can achieve realistic relighting with lighting 1 and background harmonization.}
\label{fig:ours_lighting1}
\end{figure*}

\begin{figure*}[t]
\centering
\includegraphics[width=0.88\textwidth]{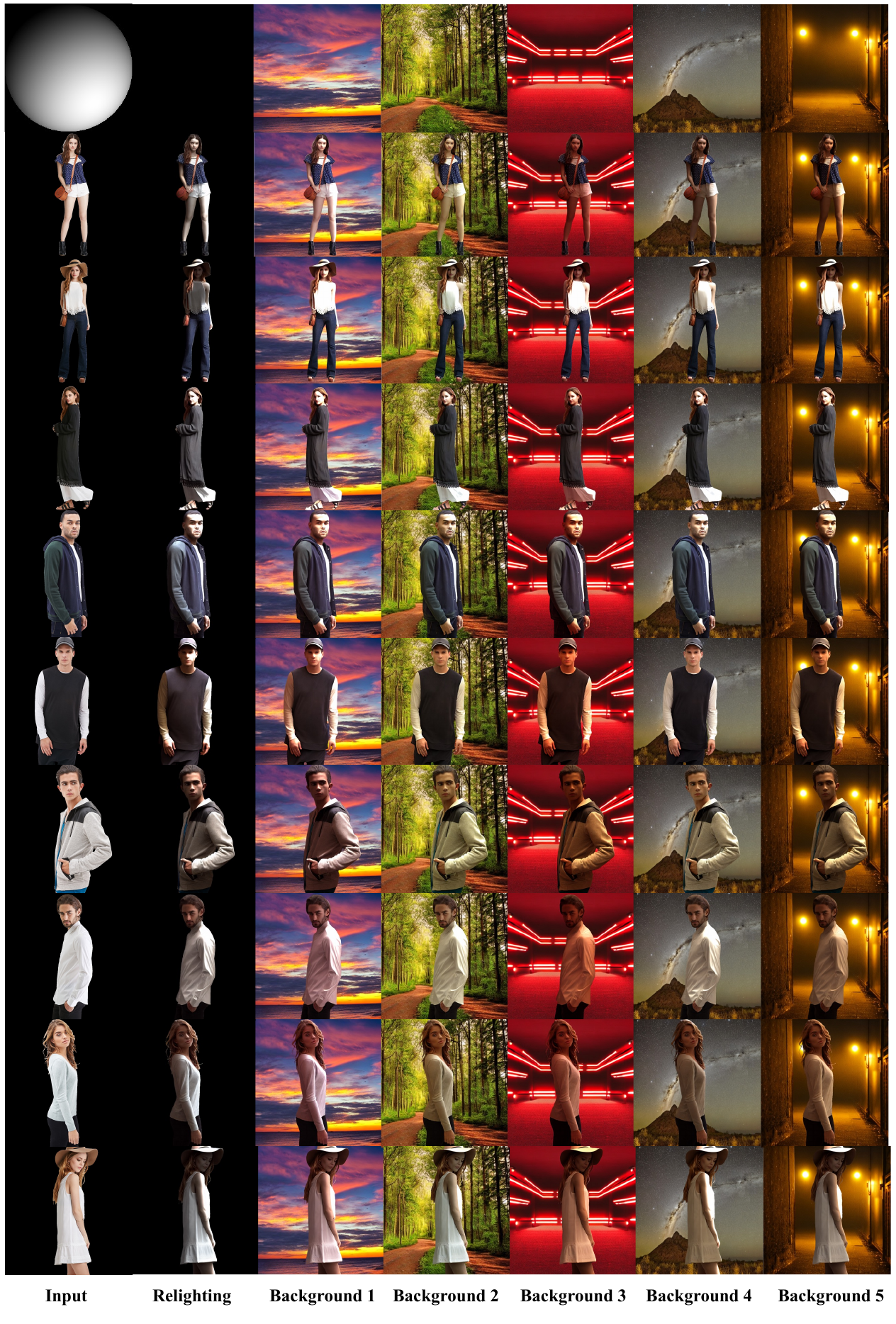}
\caption{Our model can achieve realistic relighting with lighting 2 and background harmonization.}
\label{fig:ours_lighting2}
\end{figure*}

\begin{figure*}[t]
\centering
\includegraphics[width=0.88\textwidth]{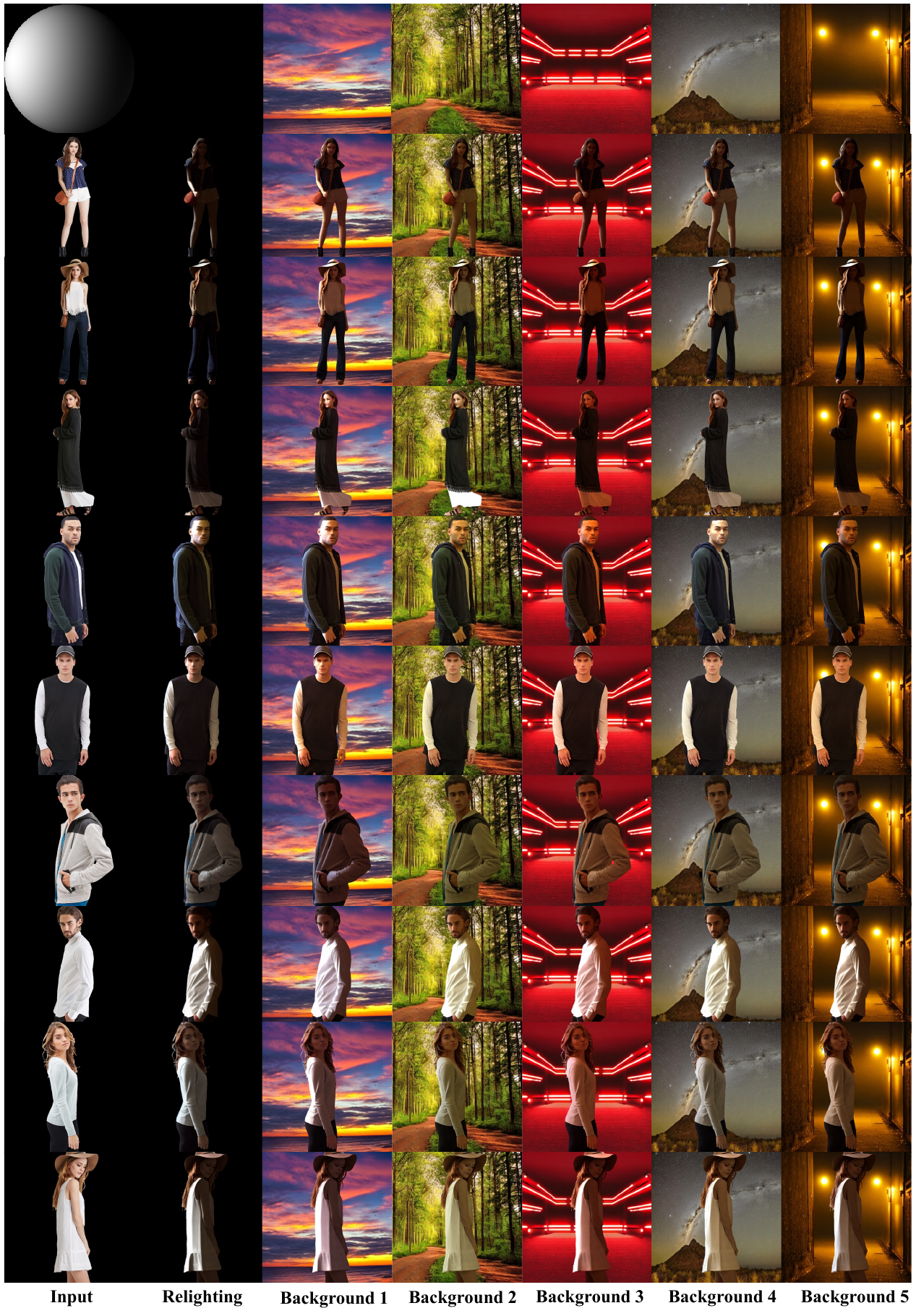}
\caption{Our model can achieve realistic relighting with lighting 3 and background harmonization.}
\label{fig:ours_lighting3}
\end{figure*}


\end{appendices}



 \fi

\end{document}